%% file: paper.tex
\PassOptionsToPackage{table,prologue,dvipsnames}{xcolor}
\documentclass[]{bytedance_seed}

\usepackage[toc,page,header]{appendix}

\newcommand{\method}{FlexWorld\xspace}

\usepackage{minitoc}
\usepackage{amsmath}
\usepackage{amssymb}

\usepackage[table]{xcolor}

\usepackage{ulem}
\usepackage{booktabs}  
\usepackage{arydshln}
\usepackage{multirow}
\usepackage{subcaption}
\usepackage{makecell}
\usepackage{xspace}
\usepackage{url}

\title{FlexWorld: Progressively Expanding 3D Scenes for Flexiable-View Synthesis}

\author[1,2,*]{Luxi Chen}
\author[1,2,*]{Zihan Zhou}
\author[3]{Min Zhao}
\author[4,\dagger]{Yikai Wang}
\author[5]{Ge Zhang}
\author[5]{Wenhao Huang}
\author[1]{Hao Sun}
\author[1]{Ji-Rong Wen}
\author[1,2,\dagger]{Chongxuan Li}

\affiliation[1]{Gaoling School of AI, Renmin University of China}
\affiliation[2]{Beijing Key Laboratory of Big Data Management and Analysis Methods}
\affiliation[3]{Dept. of Comp. Sci. \& Tech., BNRist Center, THU-Bosch MLCenter, Tsinghua University}
\affiliation[4]{School of Artificial Intelligence, Beijing Normal University}
\affiliation[5]{ByteDance Seed}

\contribution[*]{Equal Contribution}
\contribution[\dagger]{Corresponding authors}

\abstract{
Generating flexible-view 3D scenes, including 360° rotation and zooming, from single images is challenging due to a lack of 3D data. To this end, we introduce FlexWorld, a novel framework consisting of two key components: (1) a strong video-to-video (V2V) diffusion model to generate high-quality novel view images from incomplete input rendered from a coarse scene, and (2) a progressive expansion process to construct a complete 3D scene. In particular, leveraging an advanced pre-trained video model and accurate depth-estimated training pairs, our V2V model can generate novel views under large camera pose variations. Building upon it, FlexWorld progressively generates new 3D content and integrates it into the global scene through geometry-aware scene fusion. Extensive experiments demonstrate the effectiveness of FlexWorld in generating high-quality novel view videos and flexible-view 3D scenes from single images, achieving superior visual quality under multiple popular metrics and datasets compared to existing state-of-the-art methods. Qualitatively, we highlight that FlexWorld can generate high-fidelity scenes with flexible views like 360° rotations and zooming. Project page: \url{https://ml-gsai.github.io/FlexWorld}.}

\begin{document}
\maketitle


\renewcommand\cellset{\renewcommand\arraystretch{1}}
\begin{figure*}[t]
    \centering
\resizebox{1.0\textwidth}{!}{
    \begin{tabular}{cc}
    \multicolumn{1}{c|}{ 
    \begin{minipage}{0.2\textwidth}
        \centering
        \includegraphics[width=\textwidth]{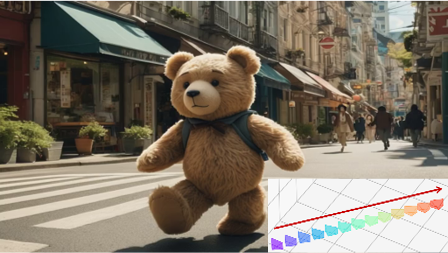}
    \end{minipage}}
    &
    \begin{minipage}{0.8\textwidth}
        \centering
        \includegraphics[width=\textwidth]{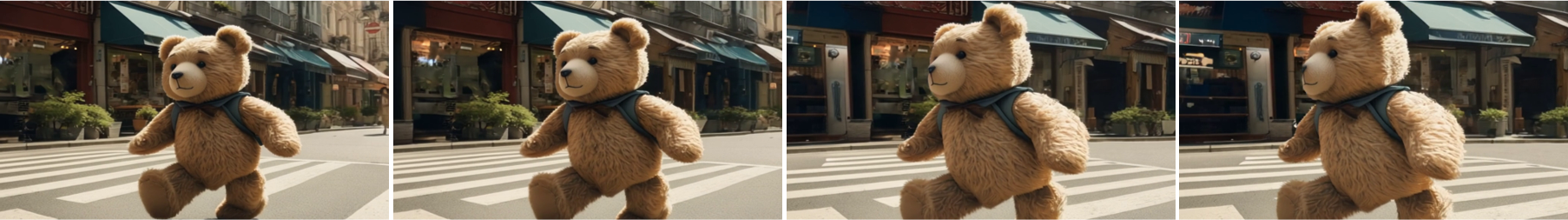}
    \end{minipage} \\
    \multicolumn{1}{c|}{
    \begin{minipage}{0.2\textwidth}
        \centering
        \includegraphics[width=\textwidth]{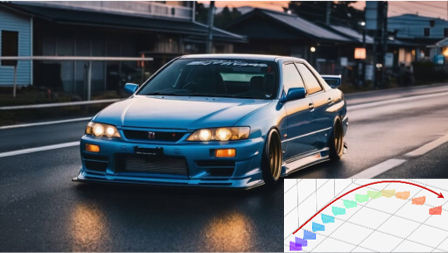}
    \end{minipage}}
    &
    \begin{minipage}{0.8\textwidth}
        \centering
        \includegraphics[width=\textwidth]{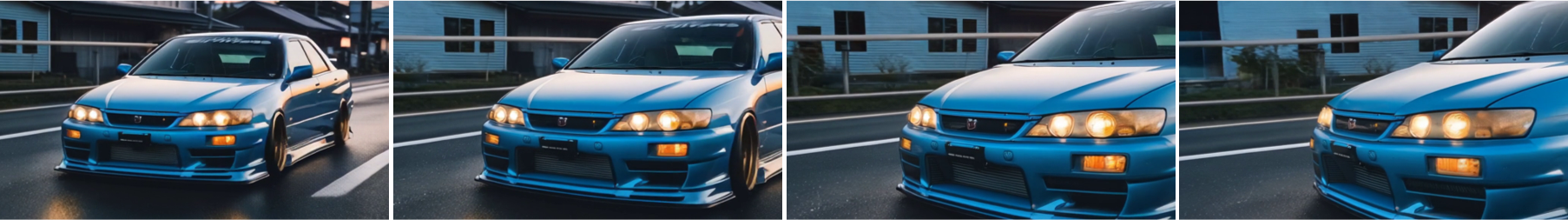}
    \end{minipage} \\
    \multicolumn{1}{c|}{
    \begin{minipage}{0.2\textwidth}
        \centering
        \includegraphics[width=\textwidth]{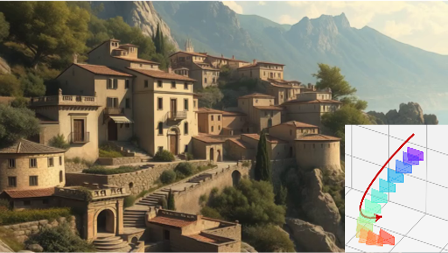}  
    \end{minipage}} 
    &
    \begin{minipage}{0.8\textwidth}
        \centering
        \includegraphics[width=\textwidth]{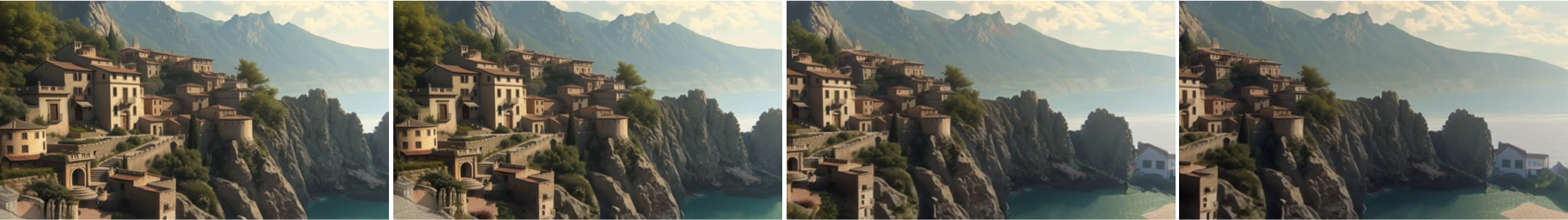} 
    \end{minipage}\vspace{0.8ex}\\ 
    (a) Input &  Videos generated by our V2V model given camera trajectories \vspace{0.8ex}\\
    \multicolumn{1}{c|}{\multirow{2}{*}{
    \begin{minipage}{0.2\textwidth}
        \centering
        \includegraphics[width=\textwidth]{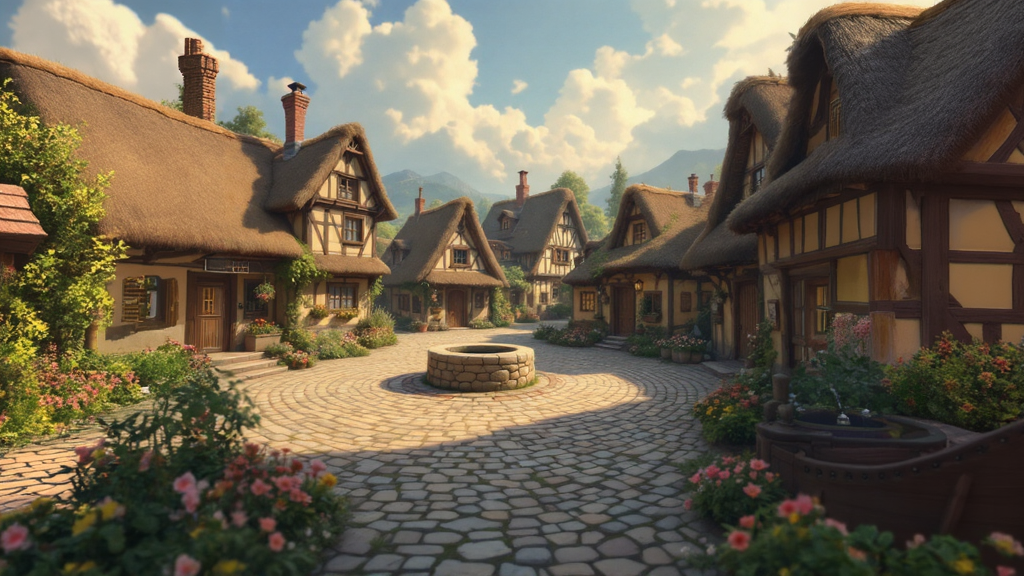}
    \end{minipage}}}
    &
    \begin{minipage}{0.8\textwidth}
        \centering
        \includegraphics[width=1.001\textwidth]{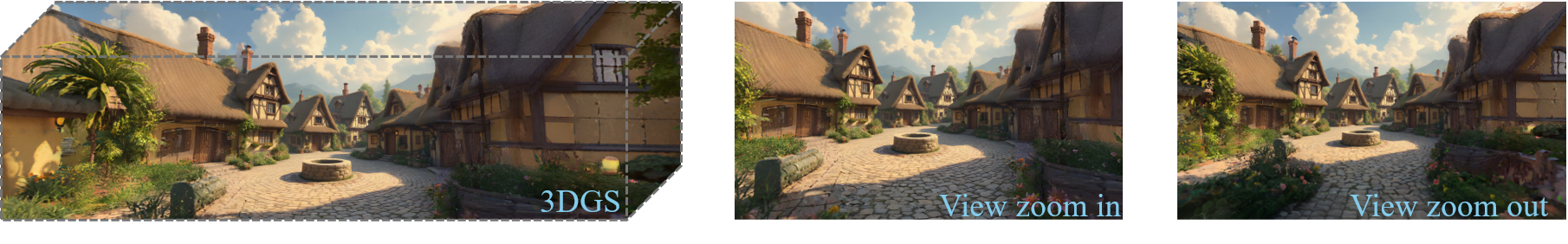}
    \end{minipage}\\
    \multicolumn{1}{c|}{}
    &
    \begin{minipage}{0.8\textwidth}
        \centering
        \includegraphics[width=\textwidth]{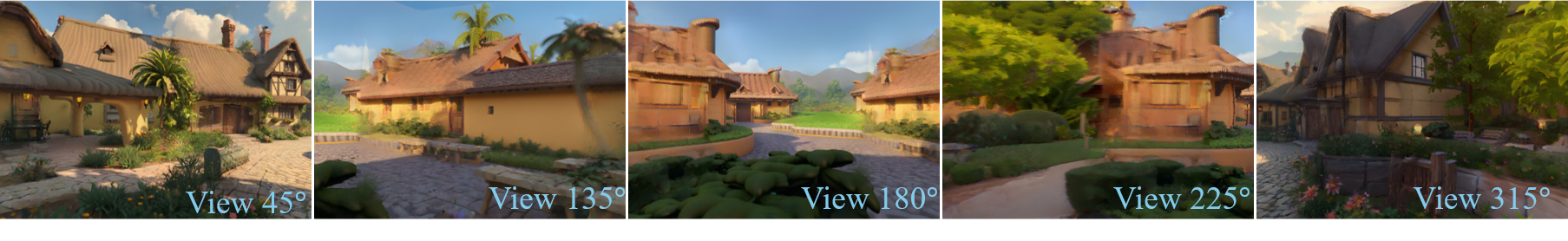}
    \end{minipage}\\
    \multicolumn{1}{c|}{\multirow{2}{*}{
    \begin{minipage}{0.2\textwidth}
        \centering
        \includegraphics[width=\textwidth]{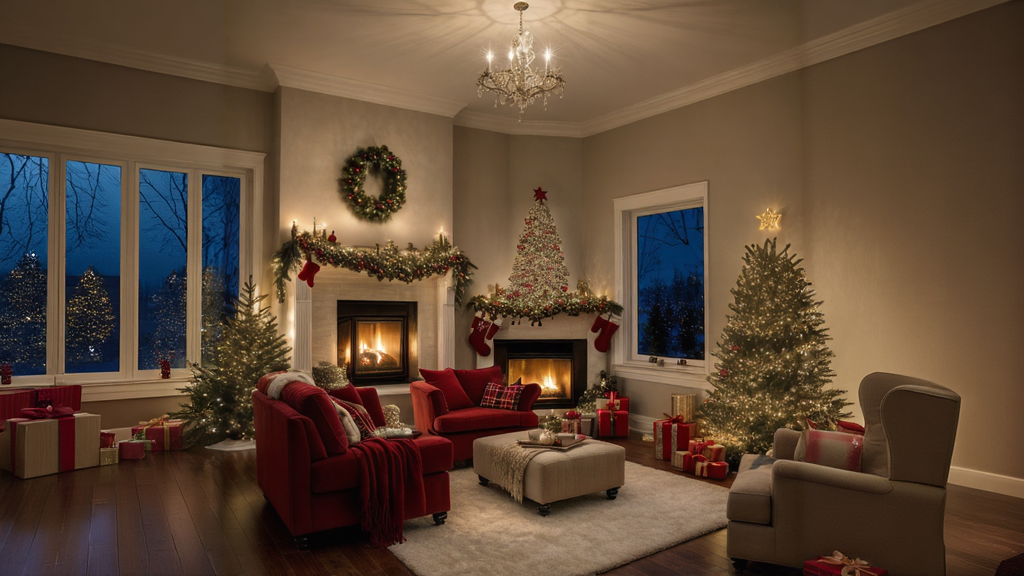}
    \end{minipage}}}
    &
    \begin{minipage}{0.8\textwidth}
        \centering
        \includegraphics[width=1.001\textwidth]{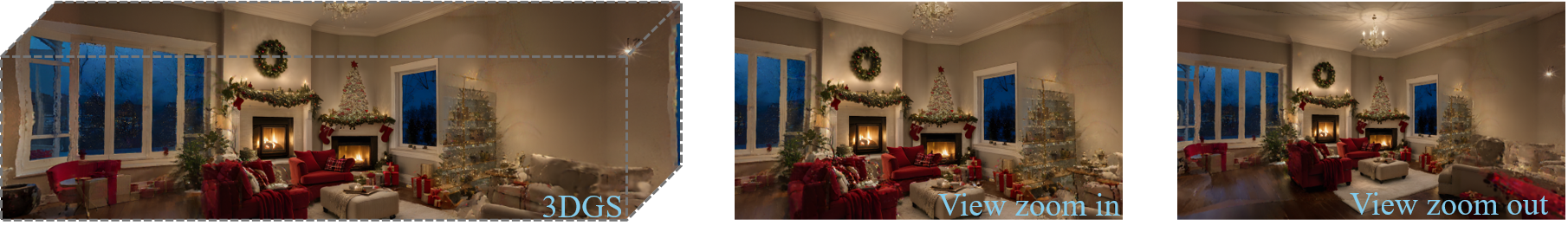}
    \end{minipage}\\
    \multicolumn{1}{c|}{}
    &
    \begin{minipage}{0.8\textwidth}
        \centering
        \includegraphics[width=\textwidth]{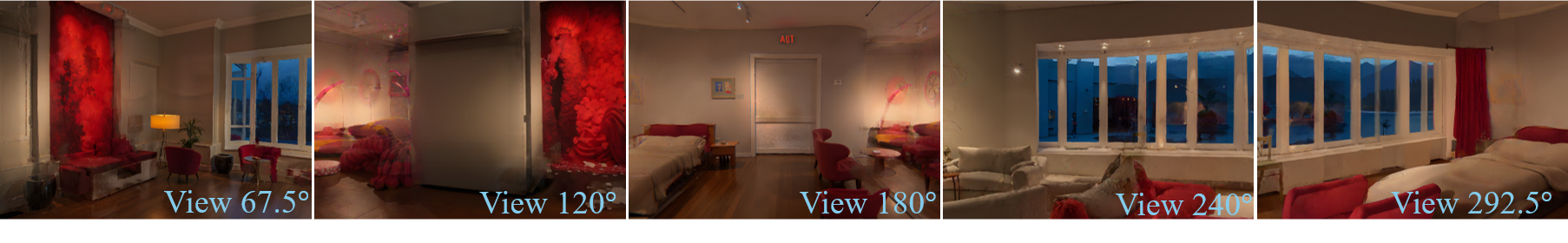}
    \end{minipage}\\
    (b) Input &  Flexible-view 3D scene generated by \method\\
    \end{tabular}
    }
\caption{\textbf{\method generates high-quality videos with camera control and flexible-view 3D scenes progressively.} (a) \method introduces a V2V diffusion producing high-quality videos from incomplete scene renderings given diverse camera trajectories with \textit{large variation}. (b) \method progressively generates \textit{flexible-views} (e.g., 360° rotations and zooming) 3DGS scenes via the V2V diffusion.} 
\vspace{-.2cm}
\label{fig:showbeforeintro}
\end{figure*}

\section{Introduction}
\label{sec:intro}

Creating a 3D scene with flexible views from a single image holds transformative potential for applications where direct 3D data acquisition is costly or impractical, such as archaeological preservation and autonomous navigation. However, this task remains fundamentally ill-posed: a single 2D observation provides insufficient information to disambiguate the complete 3D structure. In particular, when extrapolating to extreme viewpoints (e.g., 180° rotations), previously occluded or entirely absent content may emerge, introducing significant \textit{uncertainty}. 

Generative models, particularly diffusion models~\cite{sohl2015deep,ho2020denoising,song2021scorebased}, offer a principled and effective solution to this problem. While existing methods often rely on pre-trained generative models as priors for novel view synthesis, they face notable limitations. Image-based diffusion methods~\cite{lucid,wang2024vistadream,yu2024wonderjourney,yu2024wonderworld} tend to accumulate geometric errors, whereas video-based diffusion approaches~\cite{wang2024motionctrl,he2024cameractrl} struggle with dynamic content and poor camera supervision. Recent attempts~\cite{viewcrafter,ma2024you} to incorporate point cloud priors for improved consistency have shown promise but remain limited in scalability, often failing under large viewpoint changes.

To this end, we propose \method for flexible-view 3D scene generation from single images.  In contrast to existing methods~\cite{liang2024wonderland,sun2024dimensionx,zhai2025stargen}, \method \textit{progressively expands} a persistent 3D representation by synthesizing and integrating novel 3D content.  \method consists of two key components: (1) a strong video-to-video (V2V) diffusion model to generate complete view images from incomplete ones rendered from coarse scenes, and (2) a geometry-aware 3D scene expansion process, which extracts new 3D content and integrates it to the global structure. 
In particular, we fine-tuned an advanced video foundation model~\cite{yang2024cogvideox}
on accurate depth-estimated training pairs so that it can generate high-quality content under large camera variations. Built upon the V2V model, the scene expansion process employs camera trajectory planning, scene integration, and a refinement process to progressively construct a detailed 3D scene from a single image. 

Our extensive experiments show the effectiveness of 
\method in both high-quality video and flexible-view 3D scene synthesis. In particular, our V2V model achieves superior visual quality compared to the current state-of-the-art baselines~\cite{wang2024motionctrl,he2024cameractrl,sun2024dimensionx,ma2024you,viewcrafter} while maintaining excellent camera controllability across multiple benchmarks \cite{realestate10k,tanktemple}  
(see Tab.~\ref{tab:quan-comp-novel}). A similar conclusion holds for the 3D scene generation benchmarks (see Tab.~\ref{tab:quan-comp-scene}).
In addition, \method enables the synthesis of 3D scenes with a flexible view in high-fidelity (see Fig.~\ref{fig:showbeforeintro}), consistent with our quantitative results.

In summary, our key contributions are:
\begin{itemize}
    \item We introduce \method, a progressive framework for flexible-view scene generation that utilizes consistent video sequences to expand and construct scenes.
    \item We present a video-to-video diffusion model fine-tuned on carefully designed training data, enabling the generation of novel views under large camera variations while ensuring consistency with the existing scene.
    \item \method exhibits superior performance in video and scene generation compared with baseline models on various benchmark datasets~\cite{realestate10k,tanktemple}. 
\end{itemize} 

\section{Related work}

\begin{figure*}[t]
    \centering
    \includegraphics[width=.95\linewidth]{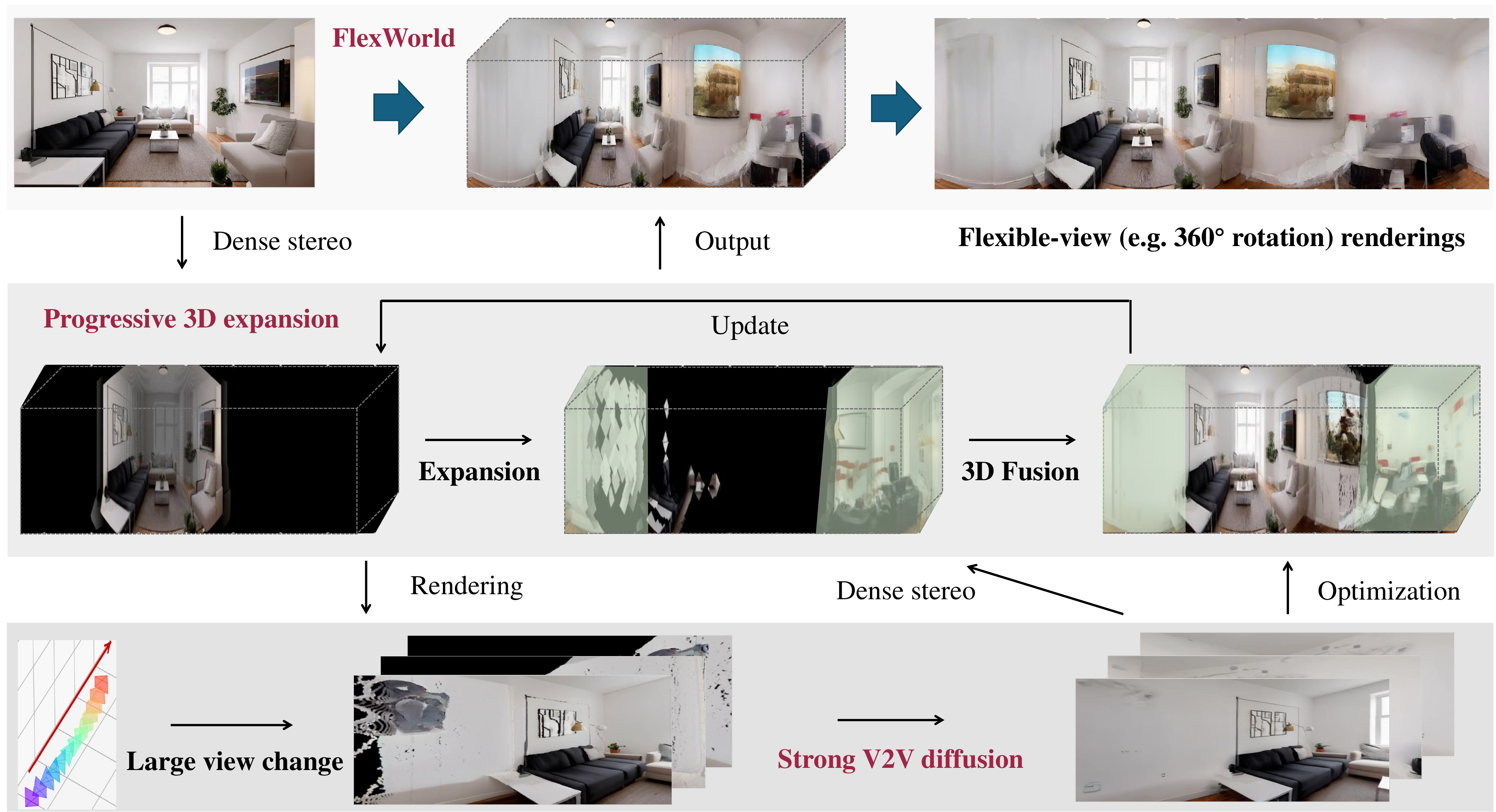} 
    \caption{\textbf{Overview of \method.}  
    \method trains a strong V2V diffusion capable of generating high-quality videos from incomplete views rendered from coarse 3D scenes. It progressively expands the 3D scene by adding new 3D content estimated from the refined videos via a dense stereo model. Ultimately, from a single image, it yields a detailed 3D scene capable of rendering flexible viewpoints.
    }  
    \label{fig:method}
\end{figure*}

\subsection{3D scene generation}
With the emergence of 3D representations that enable differentiable rendering~\cite{mildenhall2021nerf,kerbl20233d}, 3D object generation from single texts/images has advanced rapidly~\cite{poole2022dreamfusion,wang2024prolificdreamer,hong2023lrm,li2023instant3d,xu2024instantmesh,tang2023dreamgaussian,chen2024microdreamer,xiang2024structured,tang2024lgm}, closely followed by advancements in 3D scene generation. Several works~\cite{lucid,realdreamer,yu2024wonderjourney,yu2024wonderworld,wang2024vistadream,zhang2024text2nerf,hollein2023text2room} employ image diffusion models~\cite{rombach2022high,podell2023sdxl} for novel view synthesis and monocular depth estimation~\cite{bhat2023zoedepth,ranftl2020towards,ke2024repurposing,bochkovskii2024depth} to derive 3D structures for corresponding views. Another line of the work~\cite{szymanowicz2024flash3d,li2025director3d,yang2024prometheus,gslrm2024,fan2024instantsplat,chen2024mvsplat,pixelsplats} involves training a network to obtain a 3D representation from single or sparse images directly. Recent studies have integrated camera control into image~\cite{zeronvs,wu2023reconfusion,cat3d,ma2024you} or video models~\cite{liang2024wonderland,zhai2025stargen,viewcrafter,sun2024dimensionx,reconx} to facilitate the generation of novel views, subsequently performing 3D reconstruction~\cite{wang2024dust3r,leroy2024grounding} to obtain representations of 3D scenes. However, these models often struggle to generate novel views under significant view change, which limits the flexibility in generating scene viewpoints. By fine-tuning a V2V model on carefully designed training data, \method can generate novel views under significant variations in camera trajectories, thereby facilitating the creation of flexible-view 3D scenes.

\subsection{Camera-controlled video diffusion models}
\begin{figure*}[t]
    \centering
    \resizebox{1.0\textwidth}{!}{
    \begin{tabular}{cc}
    
    \begin{minipage}{0.01\textwidth}
        \centering
        \small \rotatebox{90}{ViewCrafter~\cite{viewcrafter}}
    \end{minipage}
    &
    \begin{minipage}{1.0\textwidth}
        \centering
        \includegraphics[width=\textwidth]{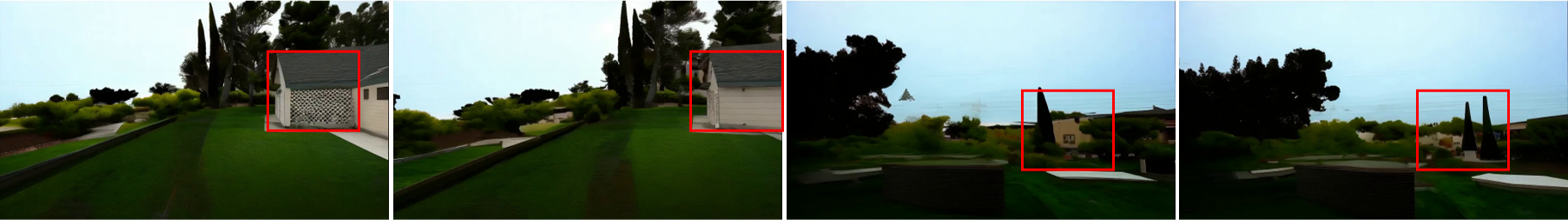}
    \end{minipage} \\
    \begin{minipage}{0.01\textwidth}
        \centering
        \small \rotatebox{90}{See3D~\cite{ma2024you}}
    \end{minipage}
    &
    \begin{minipage}{1.0\textwidth}
        \centering
        \includegraphics[width=\textwidth]{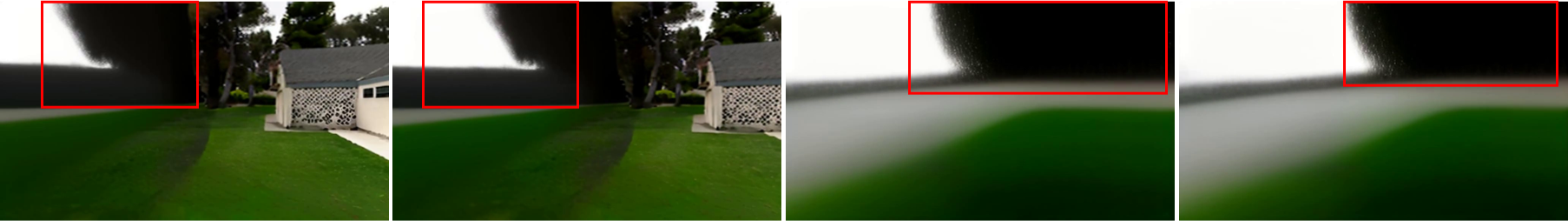}
    \end{minipage} \\
    \begin{minipage}{0.01\textwidth}
        \centering
        \small \rotatebox{90}{\textbf{Ours}}
    \end{minipage}
    &
    \begin{minipage}{1.0\textwidth}
        \centering
        \includegraphics[width=\textwidth]{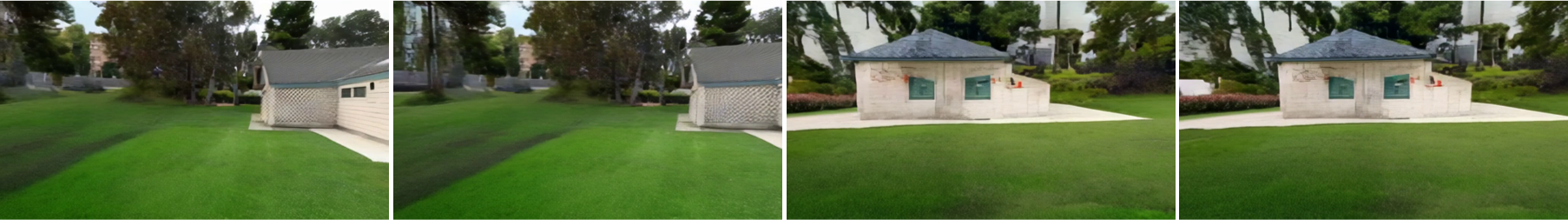}
    \end{minipage}\vspace{0.8ex} \\
    &  \makecell[cc]{
    \makebox[0.2475\linewidth]{View $45.0^\circ$} \hfill
    \makebox[0.2475\linewidth]{View $52.5^\circ$} \hfill
    \makebox[0.2475\linewidth]{View $165.0^\circ$} \hfill
    \makebox[0.2475\linewidth]{View $172.5^\circ$}
  }\\
    \end{tabular}
    }
    
\caption{\textbf{We improve our video diffusion model to enable generating 3D consistent videos under large camera variation.} We present novel views generated from each model when the camera is rotated 180 degrees to the left. The red bounding box indicates 3D inconsistency or poor visual quality in the generated content. Our model generates higher quality and more consistent static 3D scenes.} 
\vspace{-.2cm}
\label{fig:bigposechange}
\end{figure*}

Recently, Camera-controlled video diffusion models have received widespread attention. Several works~\cite{wang2024motionctrl,he2024cameractrl,bahmani2024vd3d,bahmani2024ac3d,xu2024camco} explore the generation of videos under camera conditions. However, these models are not designed for static scene generation, as the dynamics in the generated videos hinder reconstruction. DimensionX~\cite{sun2024dimensionx} achieves basic camera control via several LoRAs~\cite{hu2021lora} but lacks flexibility in complex movements. Wonderland~\cite{liang2024wonderland} and StarGen~\cite{zhai2025stargen} can generate videos from single views and camera trajectories; however, they are unable to produce new videos to complement existing 3D structures,  restricting the range of generated scenes. See3D~\cite{ma2024you} and ViewCrafter~\cite{viewcrafter} can accept missing information from specific scene perspectives and perform completion, but they struggle to accommodate significant perspective changes (see Fig.~\ref{fig:bigposechange}). In contrast, we propose training a V2V model on a more advanced video foundation model, leveraging existing scene information to enable large camera variation and offering a powerful tool for flexible view scene generation.


\section{Method}
\label{sec:method}

In this section, we will first introduce the preliminary for \method. Subsequently, we will present our flexible-view 3D scene generation framework in Sec.~\ref{sec:3d-scene-gen}. Finally, we will discuss our improved V2V model that supports our framework in Sec.~\ref{sec:videodiffusion}.

\renewcommand\cellset{\renewcommand\arraystretch{1}}

\subsection{Preliminaries}
\label{sec:preliminary}
\textbf{Video diffusion model.}
A diffusion model~\cite{sohl2015deep,ho2020denoising,song2021scorebased} consists of a forward and a denoising process. In the forward process, the diffusion model gradually adds Gaussian noise to a clean image \( x_0\) from time 0 to \( T \). The noisy image $x_t$ at a certain time \( t \in [0, T] \) can be expressed as \( x_t = \alpha_t x_0 + \sigma_t \epsilon \), where \( \alpha_t \) and \( \sigma_t \) are predefined hyperparameters. In the denoising process, a noise predictor \( \epsilon_{\theta}(x_t, t) \) with parameters \( \theta \) is trained to predict noise in \( x_t \) for generation. Given the corresponding condition $y$ for $x$, the training objective of a diffusion model is:
\begin{align}
    \min_{\theta} \mathbb{E}_{t\sim\mathcal{U}(0,1), \epsilon\sim \mathcal{N}(0, {I})}\left[\|\epsilon_{\theta}({x}_t, t; {y})-\epsilon\|_2^2\right]. \label{eq:diff-loss}
\end{align}
Recent video diffusion models~\cite{AnimateDiff,blattmann2023stable,liu2024sora,yang2024cogvideox,kong2024hunyuanvideo,zhao2025riflex} typically employ a 3D-VAE~\cite{kingma2014auto} encoder $\mathcal{E}$ to compress the source video into a latent space where the diffusion model is trained. The generated latent video is subsequently decoded to the pixel space using the corresponding decoder $\mathcal{D}$. 

\textbf{Dense stereo model.}
The dense stereo models~\cite{wang2024dust3r,leroy2024grounding,tang2024mv}, e.g., DUSt3R~\cite{wang2024dust3r} and MASt3R~\cite{leroy2024grounding}, provide an advanced tool for obtaining corresponding point maps, depth maps, and camera parameters from single or sparse views, facilitating the reconstruction of 3D point clouds. This approach offers a means to derive a rough 3D structure and camera estimation from a single image.

\textbf{3D Gaussian splatting.} As currently one of the most popular 3D representations, 
3D Gaussian splatting (3DGS)~\cite{kerbl20233d} models the 3D scene by multiple 3D Gaussians parameterized by colors, centers, opacities, scales, and rotation quaternions. The effectiveness and efficiency of 3DGS in 3D reconstruction and generation have been widely demonstrated~\cite{kerbl20233d,pixelsplats,chen2024mvsplat,tang2023dreamgaussian,chen2024microdreamer,tang2024lgm}. In addition to the $\mathcal{L}_{1}$ loss and SSIM loss $\mathcal{L}_{\text{SSIM}}$~\cite{ssim} presented in the original paper~\cite{kerbl20233d}, optimizing a 3D scene's loss function typically incorporates the LPIPS loss $\mathcal{L}_{\text{LPIPS}}$~\cite{lpips,sun2024dimensionx} to  improve optimization. The weights \(\lambda_1\), \(\lambda_{\text{SSIM}}\), and \(\lambda_{\text{LPIPS}}\) are adjustable hyperparameters. Formally, the specific loss function is expressed as:
\begin{align}
    \mathcal{L}=\lambda_1\mathcal{L}_{1}+\lambda_{\text{SSIM}}\mathcal{L}_{\text{SSIM}}+\lambda_{\text{LPIPS}}\mathcal{L}_{\text{LPIPS}}. \label{eq:loss-3d}
\end{align}

\subsection{Progressive flexible-view 3D scene expansion}
\label{sec:3d-scene-gen}

To overcome the limitation of insufficient multi-views in single videos for 3D scene generation discussed in Sec.~\ref{sec:intro}, we propose a progressive scene expansion method, named \method. \method consists of two key parts: 1) novel view video synthesis guided by pre-defined camera trajectories that explore unseen regions and 2) geometry-aware 3D scene expansion that updates a persistent global structure while maintaining geometric coherence. We employ a video-to-video (V2V) model that can generate corresponding high-quality videos based on incomplete videos rendered from a rough scene for novel view synthesis, which will be discussed in detail in Sec.~\ref{sec:videodiffusion}.

In this section, we focus on geometry-aware 3D scene expansion, which consists of three key subparts: (1) \textit{Camera trajectory planning}, which determines the regions to be expanded; (2) \textit{Scene integration}, which integrates the newly generated 3D content into the global scene; and (3) \textit{Refinement process}, which enhances the visual quality of the scene. Each component is discussed in detail below.


\textbf{Camera trajectory planning.} Camera trajectory determines the regions to be expanded, but areas without \textit{any} 3D information will lose camera control for V2V model. We prioritize camera movement toward specific regions to ensure the input videos always contain 3D information. Specifically, starting with an initial rough scene from a single image, we first generate novel views by zooming out to expand the scene. Next, we alternately rotate the camera 180° left and right to add more scene details, ultimately achieving a flexible 360° view. See Sec.~\ref{sec:abla} for ablation on the necessity of the initial trajectory of zooming out.

\textbf{Scene intergration.} To extract 3D information from the generated video, we first develop a method to obtain new 3D content, which is then integrated into the global scene. Subsequently, we treat all video frames as ground truth and optimize the scene to enhance reconstruction.

Formally, we select $m$ keyframes from the generated video to facilitate the extraction of 3D content, i.e., point cloud. We utilize DUSt3R~\cite{wang2024dust3r} to generate initial depth maps $\hat{D}_0,...,\hat{D}_m$ for each of the \(m\) keyframes $I_1,...,I_{m}$ and a reference view $I_0$ simultaneously. For each view, we render the corresponding incomplete depth maps $D_0,..., D_m$ from the existing scene, along with their masks $M_0,...,M_m$. The reference view is usually well optimized, and its rendered depth $D_0$ is completely known and can be used to measure the depth scale. For each $1\leq i\leq m$, the new adding point cloud $\mathcal{P}_i$ from view $i$ can be obtained by:
\begin{align}
    \tilde{D}_i&=\text{Depth-align}\left(\frac{\text{Median}(D_0))}{\text{Median}(\hat{D}_0))}\cdot\hat{D}_i,D_i,M_i\right),\\
    \mathcal{P}_i&=\{\tilde{D}_i(u,v) E_i^{-1}K^{-1}\cdot (u,v,1)^T|M_i(u,v)=1\},
\end{align}
where $E_i$ denotes extrinsic for keyframe $i$, $K$ denotes intrinsic, and $(u,v)$ stands for the pixel coordinates of the frame, ranging from $0$ to frame size. $\text{Median}(\cdot)$ represents extracting the median value from given depth map. By aligning the depth scale of the reference view, we mitigate the instability inherent in the depth estimation model. $\text{Depth-align}(\cdot)$ denotes any further depth alignment operation, and we employ guided filtering~\cite{he2012guided} here to achieve smoother overall depth transitions. In the end, we convert these point clouds $\{\mathcal{P}_1,..., \mathcal{P}_m\}$ to 3DGS and add to the scene and optimize using views from the entire video with the loss function in Eq.~(\ref{eq:loss-3d}). The corresponding hyperparameters can be seen in Sec.~\ref{sec:expr}.

\textbf{Refine process.}
To further enhance the visual quality of the generated scene, we adopt SDEdit~\cite{meng2021sdedit} by rendering multi-view images ${I}$ from fixed viewpoints, adding random noise, and applying a multi-step denoising process using the FLUX.1-dev~\cite{flux_repo} image diffusion model:
\begin{align}
    \hat{I}=f_{\theta}(\alpha_{t}I+\sigma_t\epsilon, t),
\end{align}
where \( t \) is the timestamp of the forward diffusion process, \( f_{\theta} \) represents the denoising model, and \( \hat{I} \) denotes the refined images. We use \( \hat{I} \) to refine the corresponding views of the 3DGS scene with the same loss function as Eq.~(\ref{eq:loss-3d}). See Fig.~\ref{fig:abla} for ablation.

From the discussion above, we note that while the proposed framework is feasible, generating flexible-view scenes requires the V2V model to produce new consistent content under large camera variation. Otherwise, achieving flexible-view scenes may necessitate multiple iterations, potentially introducing cumulative errors that affect the consistency of the 3D scene.

\subsection{Improved diffusion for novel view synthesis}
\label{sec:videodiffusion}
\renewcommand\cellset{\renewcommand\arraystretch{1}}
\begin{figure}[t]
    \centering
    \resizebox{1.0\linewidth}{!}{
    \begin{tabular}{cc}
    
    \begin{minipage}{0.01\linewidth}
        \centering
        \small \rotatebox{90}{Ground Truth}
    \end{minipage}
    &
    \begin{minipage}{1.0\linewidth}
        \centering
        \includegraphics[width=\linewidth]{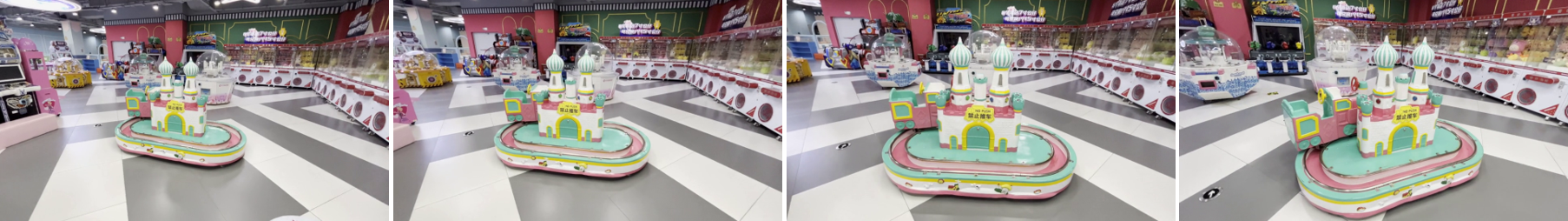}
    \end{minipage} \\
    \begin{minipage}{0.01\linewidth}
        \centering
        \small \rotatebox{90}{MASt3R~\cite{leroy2024grounding}}
    \end{minipage}
    &
    \begin{minipage}{1.0\linewidth}
        \centering
        \includegraphics[width=\linewidth]{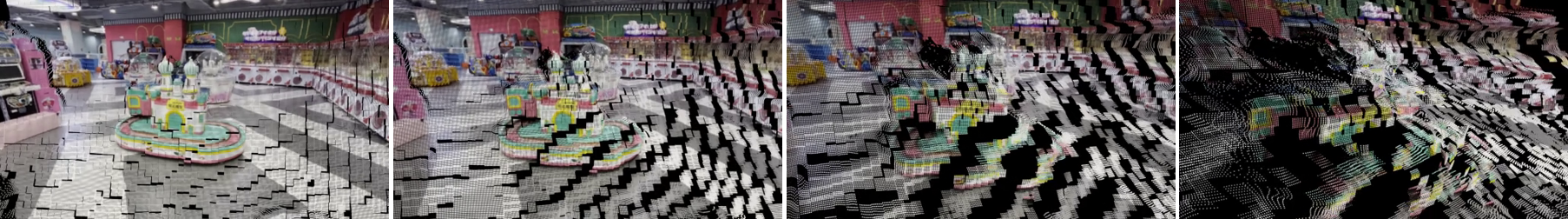}
    \end{minipage} \\
    \begin{minipage}{0.01\linewidth}
        \centering
        \small \rotatebox{90}{\textbf{Ours}}
    \end{minipage}
    &
    \begin{minipage}{1.0\linewidth}
        \centering
        \includegraphics[width=\linewidth]{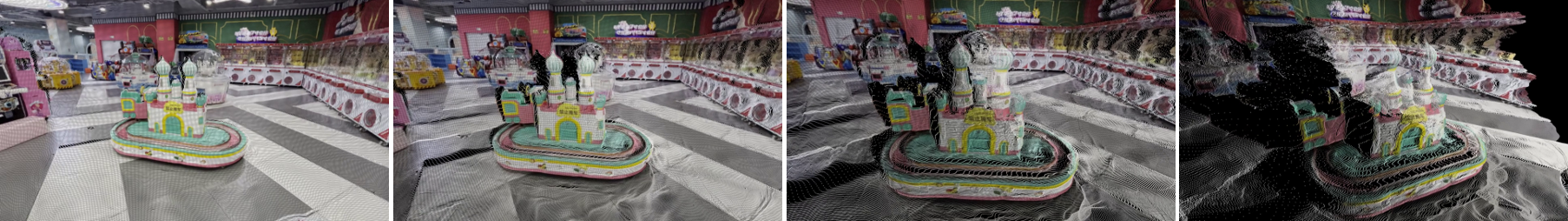}
    \end{minipage}\vspace{0.8ex} \\
    &  \makecell[cc]{
    \makebox[0.2475\linewidth]{Frame 10} \hfill
    \makebox[0.2475\linewidth]{Frame 20} \hfill
    \makebox[0.2475\linewidth]{Frame 30} \hfill
    \makebox[0.2475\linewidth]{Frame 49}
  }\\
    \end{tabular}
    }
\caption{\textbf{Our dataset construction method yields more accurate training pairs.} We present frames of incomplete videos rendered from initial point clouds generated by a dense stereo model MASt3R~\cite{leroy2024grounding} (i.e., ViewCrafter~\cite{viewcrafter}'s dataset construction method) and our 3DGS reconstruction. Our approach produces incomplete videos with better alignment to ground truth, resulting in higher-quality training pairs.} 
\vspace{-.2cm}
\label{fig:datacreate}
\end{figure}

Existing V2V approaches fail to handle significant viewpoint changes (180°)~\cite{viewcrafter,ma2024you}, as the results show in Fig.~\ref{fig:bigposechange}, primarily due to using weaker base models~\cite{blattmann2023stable,xing2024dynamicrafter} trained on suboptimal data, as shown in Fig.~\ref{fig:datacreate}. 
We improve our V2V diffusion model by conducting video conditioning on an advanced base model and carefully designed training data. 

\textbf{Video conditioning.} 
To develop a V2V diffusion model suitable for \method, we select a more advanced base model, CogVideoX-5B-I2V~\cite{yang2024cogvideox}, as our foundation and replace the original image conditioning with video conditioning. Specifically, the 3D-VAE encoder compresses the conditional videos, which are then concatenated channel-wise with noise latents. Given a camera trajectory \( {c} \), our model aims to learn the distribution \( {x} \sim p({x} | {y}) \), where \( {y} \) represents the incomplete video rendered from the rough scene under \( {c} \), and \( {x} \) denotes the high-quality video of the scene from the corresponding view. The training objective aligns with the original diffusion model, as formulated in Eq.~(\ref{eq:diff-loss}).

\textbf{Training data construction.} As illustrated in Fig~\ref{fig:datacreate}, inaccurate training pairs created by the dense stereo model~\cite{viewcrafter,wang2024dust3r,leroy2024grounding} often exhibit significant deviations from the ground truth and contain fragmented textures. As a result, the trained V2V model also demonstrates similar artifacts (see Appendix~\ref{apd:viewcrafterbad}). To address this issue, we carefully construct training data pairs by employing 3DGS reconstruction to obtain more precise depth estimation. We implement a synthetic training data pipeline as follows: 
\begin{itemize}
\item Reconstruct the 3DGS scene using all available images;
\item Start at a random frame, extract its depth from the 3DGS, and perform back-projection to obtain a point cloud;
\item Render incomplete video sequences \( {y} \) (49 consecutive frames) under complex camera trajectories from datasets;
\item Pair \( {y} \) with ground truth \( x \) to form training pairs \( ({x}, {y}) \).
\end{itemize}
This approach ensures more precise depth estimation in the generated data, resulting in a more accurate initial point cloud. This enhancement can improve the quality of training pairs, as illustrated in Fig.~\ref{fig:datacreate}.

To support the generation of static scenes and large camera variations, we select the high-quality DL3DV-10K~\cite{ling2024dl3dv} scene dataset, which contains various camera movements. We exclude the RealEstate10K dataset~\cite{realestate10k} from our training dataset, as its videos frequently contain moving objects and simple camera motions, which fail to meet our needs.

After training, our video model generalizes to generate high-quality novel views for rough scenes from incomplete inputs under arbitrary trajectories, particularly under large camera variation (see Fig.~\ref{fig:bigposechange}). This positions our model as the optimal video diffusion model in \method, significantly enhancing the generation of flexible-view 3D scenes.

\section{Experiment}
\label{sec:expr}

We present the implementation details for \method, comparison of novel view synthesis and 3D scene generation, and ablation study sequentially.

\subsection{Implementation details}
\label{sec:imple-detail}
We build our video-to-video model based on the image-conditioned video diffusion model CogVideoX-5B-I2V~\cite{yang2024cogvideox}. The model is trained at a resolution of $480 \times 720$, with a learning rate of 5e-5 and a batch size of 32, for a total of 5000 steps on 16 NVIDIA A800 80G GPUs. We retain the default settings for other hyperparameters in the original I2V fine-tuning process. In the training dataset, we utilize data from the DL3DV-10K dataset~\cite{ling2024dl3dv}, discarding any data with failed COLMAP camera annotations. The coefficient for the 3DGS loss function, specifically $\lambda_1$, $\lambda_{\text{SSIM}}$, and $\lambda_{\text{LPIPS}}$, are set to 0.8, 0.2, and 0.3, respectively. More details can be found in Appendix~\ref{apd:more-detail}.

\subsection{Comparison on novel view synthesis}
\label{sec:comp-video}
\renewcommand\cellset{\renewcommand\arraystretch{1}}
\begin{figure*}[t]
    \centering
    \resizebox{0.91\textwidth}{!}{
    \begin{tabular}{cc}
    \begin{minipage}{0.01\textwidth}
        \centering
        \small \rotatebox{90}{Input}
    \end{minipage}
    &
    \begin{minipage}{1.0\textwidth}
        \centering
        \includegraphics[width=\textwidth]{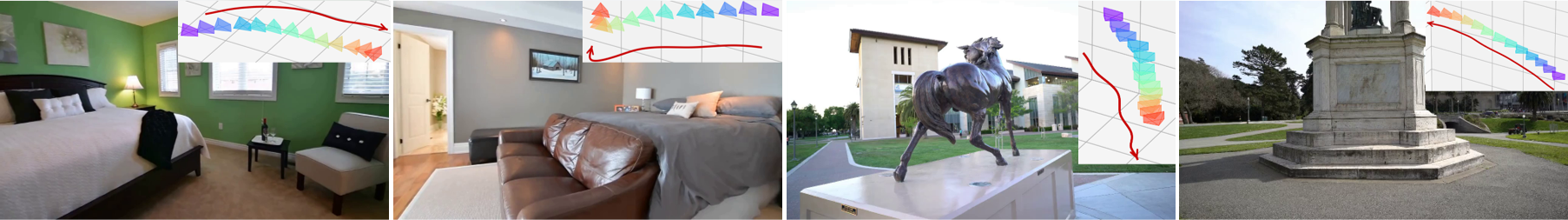}
    \end{minipage} \\
    \begin{minipage}{0.01\textwidth}
        \centering
        \small \rotatebox{90}{Ground Truth}
    \end{minipage}
    &
    \begin{minipage}{1.0\textwidth}
        \centering
        \includegraphics[width=\textwidth]{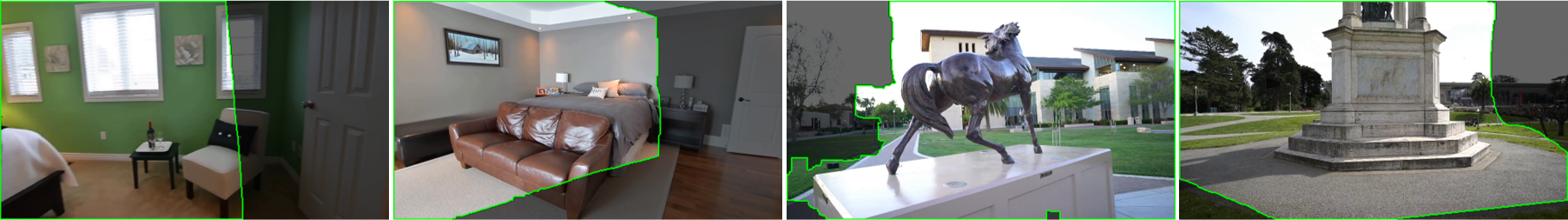}
    \end{minipage} \\
    \begin{minipage}{0.01\textwidth}
        \centering
        \small \rotatebox{90}{MotionCtrl~\cite{wang2024motionctrl}}
    \end{minipage}
    &
    \begin{minipage}{1.0\textwidth}
        \centering
        \includegraphics[width=\textwidth]{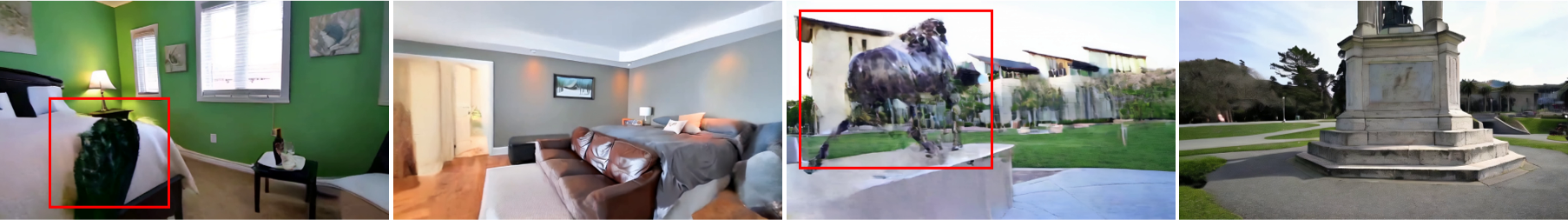}
    \end{minipage} \\
    \begin{minipage}{0.01\textwidth}
        \centering
        \small \rotatebox{90}{CameraCtrl~\cite{he2024cameractrl}}
    \end{minipage}
    &
    \begin{minipage}{1.0\textwidth}
        \centering
        \includegraphics[width=\textwidth]{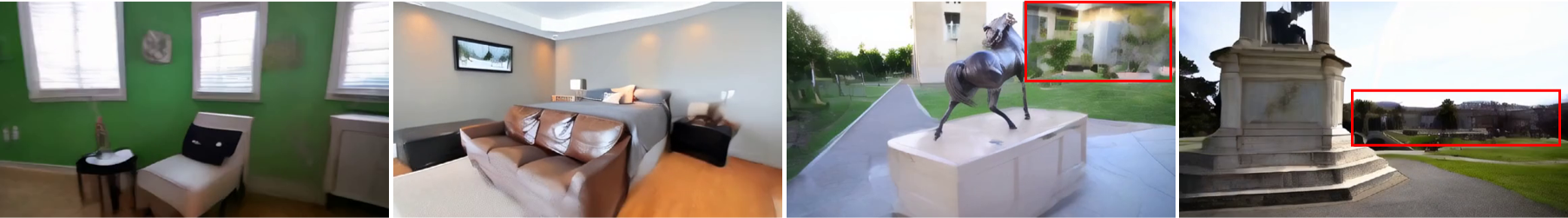}
    \end{minipage} \\
    \begin{minipage}{0.01\textwidth}
        \centering
        \small \rotatebox{90}{ViewCrafter~\cite{viewcrafter}}
    \end{minipage}
    &
    \begin{minipage}{1.0\textwidth}
        \centering
        \includegraphics[width=\textwidth]{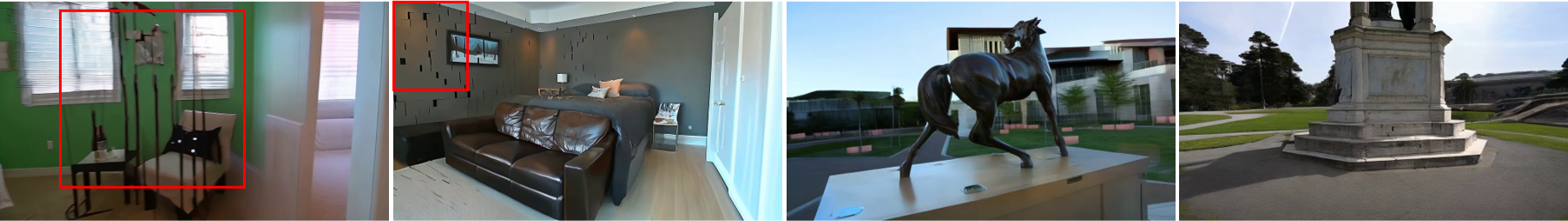}
    \end{minipage} \\
    \begin{minipage}{0.01\textwidth}
        \centering
        \small \rotatebox{90}{See3D~\cite{ma2024you}}
    \end{minipage}
    &
    \begin{minipage}{1.0\textwidth}
        \centering
        \includegraphics[width=\textwidth]{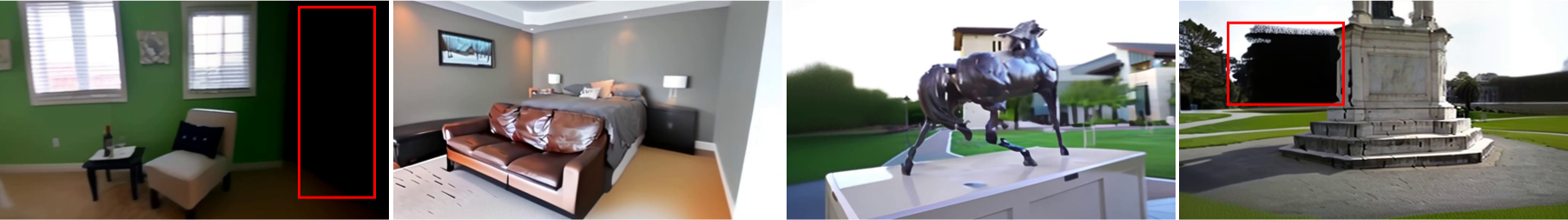}
    \end{minipage} \\
    \begin{minipage}{0.01\textwidth}
        \centering
        \small \rotatebox{90}{\textbf{Ours}}
    \end{minipage}
    &
    \begin{minipage}{1.0\textwidth}
        \centering
        \includegraphics[width=\textwidth]{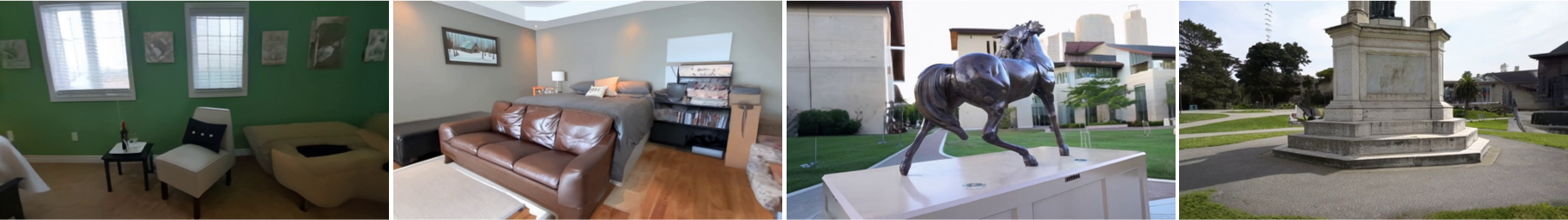}
    \end{minipage} \\
    \end{tabular}
    }
\caption{\textbf{Qualitative comparison on novel view synthesis.} We assessed the generative capabilities of various models using the same camera trajectory, focusing on the midpoint. The green bounding box in the ground truth highlights regions requiring consistency with the input, while the remaining areas demand coherent content generation. The red bounding box marks low-quality outputs in baseline models. Our model demonstrates superior visual generation quality, even under effectively controlled camera conditions.} 
\vspace{-.2cm}
\label{fig:qual-comp-novel}
\end{figure*}


\begin{table}[t]
	\centering
  \caption{\textbf{Quantitative comparison on novel view synthesis.} Our method achieves superior visual quality while maintaining commendable camera control compared to the baselines.}
  \label{tab:quan-comp-novel}
  \vspace{-.15cm}
         {
	\begin{tabular}{lccccccc}
		\toprule 
        Metric & FID $\downarrow$ & FVD $\downarrow$ & PSNR $\uparrow$ & SSIM $\uparrow$ & LPIPS $\downarrow$ & $R_{\textrm{err}}$ $\downarrow$ & $T_{\textrm{err}}$ $\downarrow$\\
        \midrule
        \multicolumn{8}{l}{\textit{RealEstate10K}} \\
        MotionCtrl & 20.41 & 226.62 & 13.19&0.516 &0.515       & 0.141 & \underline{0.216}\\
        CameraCtrl & 22.73 & 381.38 & 16.03& 0.604& 0.416      & 0.040& \textbf{0.117}\\
        DimensionX  & 33.77 & 548.19 & 11.77 & 0.491&0.659 & 0.864 & 0.615\\
        See3D & 24.24 & 259.62 & 14.44 & 0.546 & 0.477         & \textbf{0.026}  & 0.355 \\
        ViewCrafter &  \underline{16.99} & \underline{143.89}& \underline{15.74} & \underline{0.595} & \underline{0.372}   & 0.032& 0.380\\
        \cellcolor{gray!10}\textbf{\method} & \cellcolor{gray!10}\textbf{13.88} & \cellcolor{gray!10}\textbf{100.41} & \cellcolor{gray!10}\textbf{16.62} & \cellcolor{gray!10}\textbf{0.612} & \cellcolor{gray!10}\textbf{0.344} & \cellcolor{gray!10}\textbf{0.026} & \cellcolor{gray!10}0.297\\
        \midrule
        \multicolumn{8}{l}{\textit{Tanks and Temples}} \\
        MotionCtrl & 54.24 & 651.47 & 11.39 &0.361 &0.606       & 0.336& 0.589\\
        CameraCtrl & 60.21 & 1338.53 & 11.08& 0.363&0.688       & 0.202& 0.535\\
        DimensionX  & 54.13 & 1051.15 & 11.26 &0.358& 0.678 & 0.878& 0.695\\
        See3D & 53.29 & 564.19 & \underline{12.95} & \underline{0.404} & 0.584          & \textbf{0.035}& \textbf{0.108}\\
        ViewCrafter & \underline{41.18} & \underline{549.10} & 12.52 & 0.386 & \underline{0.526}    & 0.111& 0.200\\
        \cellcolor{gray!10}\textbf{\method} & \cellcolor{gray!10}\textbf{37.31} & \cellcolor{gray!10}\textbf{376.49} & \cellcolor{gray!10}\textbf{13.20} &\cellcolor{gray!10}\textbf{0.405} & \cellcolor{gray!10}\textbf{0.525}   & \cellcolor{gray!10}\underline{0.048} & \cellcolor{gray!10}\underline{0.100}\\
        \bottomrule
	\end{tabular}
        }
    \vspace{-.1in}
\end{table} 

We evaluate the capability of our video-to-video model for novel view synthesis by comparing the visual generation quality and camera accuracy of 5 open-source baseline models, including MotionCtrl~\cite{wang2024motionctrl}, CameraCtrl~\cite{he2024cameractrl}, DimensionX~\cite{sun2024dimensionx}, See3D~\cite{ma2024you}, ViewCrafter~\cite{viewcrafter}. 

\textbf{Evaluation datasets.} To ensure fairness, we selected the RealEstate10K (RE10K) test dataset~\cite{realestate10k} and Tanks-and-Temples (Tanks)~\cite{tanktemple} datasets, which are separate from our training dataset, for evaluation. Following previous work~\cite{liang2024wonderland,viewcrafter}, we randomly selected 300 video clips with a sample stride ranging from 1 to 3 in the RealEstate10K. In the Tanks-and-Temples dataset, we randomly sampled 100 video clips with a stride of 4 across 14 test scenes. Notably, this dataset does not contain pre-labeled cameras; therefore, we utilized the MASt3R~\cite{leroy2024grounding} model to annotate the cameras. Each selected video clip involves a camera length of 49. For models generating fewer than 49 frames, we uniformly excluded cameras from the original trajectory to match the required length.

\textbf{Evaluation metrics.} 
We followed previous works~\cite{liang2024wonderland,viewcrafter} to evaluate the generated videos using various metrics comprehensively. The metrics include FID~\cite{heusel2017gans} and FVD~\cite{fvd} for assessing visual quality, as well as PSNR, SSIM~\cite{ssim}, and LPIPS~\cite{lpips} to evaluate the similarity between the generated frames and the ground truth, with the average of the calculated metrics for each frame taken. Additionally, we estimated the corresponding camera poses for each generated frame and the ground truth using MASt3R~\cite{leroy2024grounding}. The camera accuracy was calculated using the formula from prior research~\cite{he2024cameractrl,viewcrafter,liang2024wonderland}.

\textbf{Qualitative comparison.} From the qualitative comparison shown in Fig.~\ref{fig:qual-comp-novel}, all models exhibit a certain level of control over camera movements, and methods like ViewCrafter, See3D, and \method demonstrated relatively precise control; however, the visual quality of the generated outputs varied. The results from MotionCtrl often exhibited artifacts, while the content produced by CameraCtrl appeared somewhat blurred. See3D struggled to generate distinct new objects from novel viewpoints, and ViewCrafter produced dark content. In contrast, our method maintained effective camera control and surpassed all baseline models in the visual quality of the generated content.

\textbf{Quantitative comparisons.} Our quantitative results are presented in Tab.~\ref{tab:quan-comp-novel}. \method outperforms all baselines across datasets, achieving the best FID and FVD scores, indicating that generated content distribution closely aligns with the ground truth. It also attains optimal PSNR, SSIM, and LPIPS scores, demonstrating superior visual quality. Additionally, our model excels in camera control, with lower $R_\text{err}$ and $T_\text{err}$ values.

\subsection{Comparison on scene generation}

\renewcommand\cellset{\renewcommand\arraystretch{1}}
\begin{figure*}[t]
    \centering
    \resizebox{0.91\textwidth}{!}{
    \begin{tabular}{cc}
    \begin{minipage}{0.01\textwidth}
        \centering
        \small \rotatebox{90}{Input}
    \end{minipage}
    &
    \begin{minipage}{1.0\textwidth}
        \centering
        \includegraphics[width=\textwidth]{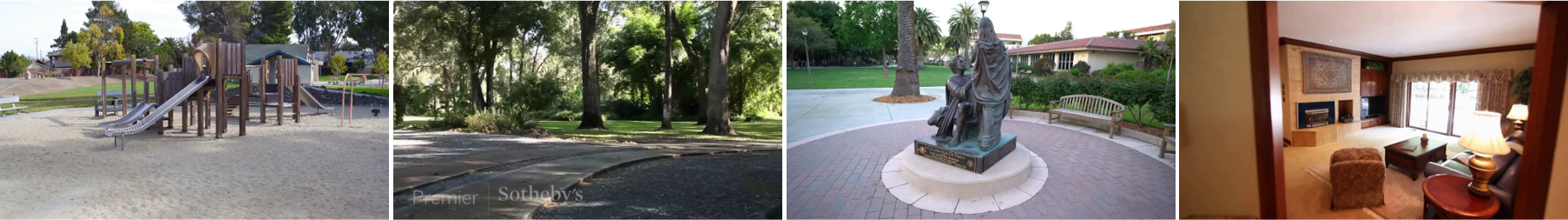}
    \end{minipage} \\
    \begin{minipage}{0.01\textwidth}
        \centering
        \small \rotatebox{90}{Ground Truth}
    \end{minipage}
    &
    \begin{minipage}{1.0\textwidth}
        \centering
        \includegraphics[width=\textwidth]{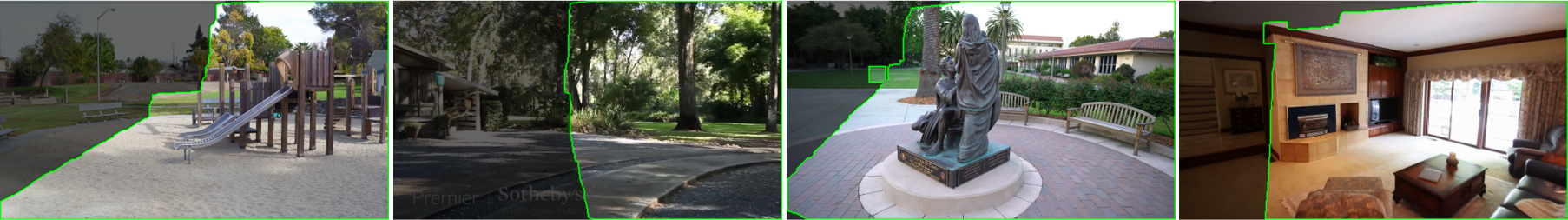}
    \end{minipage} \\
    \begin{minipage}{0.01\textwidth}
        \centering
        \small \rotatebox{90}{LucidDreamer~\cite{lucid}}
    \end{minipage}
    &
    \begin{minipage}{1.0\textwidth}
        \centering
        \includegraphics[width=\textwidth]{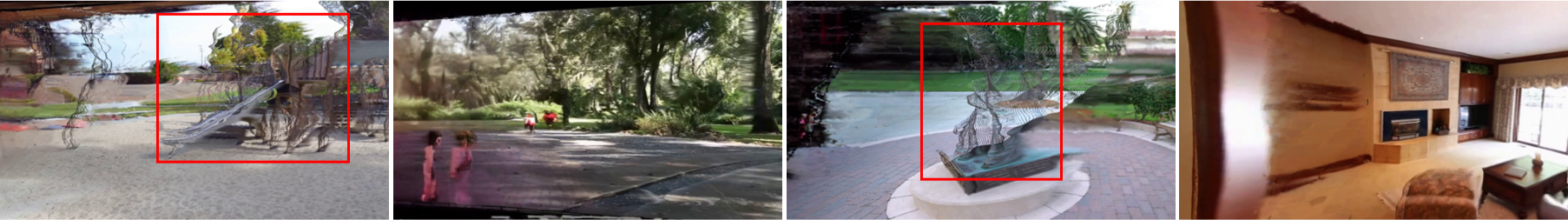}
    \end{minipage} \\
    \begin{minipage}{0.01\textwidth}
        \centering
        \small \rotatebox{90}{ViewCrafter~\cite{viewcrafter}}
    \end{minipage}
    &
    \begin{minipage}{1.0\textwidth}
        \centering
        \includegraphics[width=\textwidth]{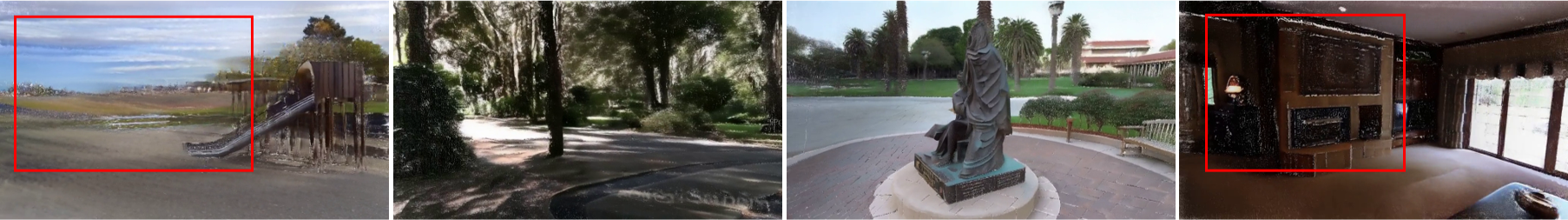}
    \end{minipage} \\
    \begin{minipage}{0.01\textwidth}
        \centering
        \small \rotatebox{90}{See3D~\cite{ma2024you}}
    \end{minipage}
    &
    \begin{minipage}{1.0\textwidth}
        \centering
        \includegraphics[width=\textwidth]{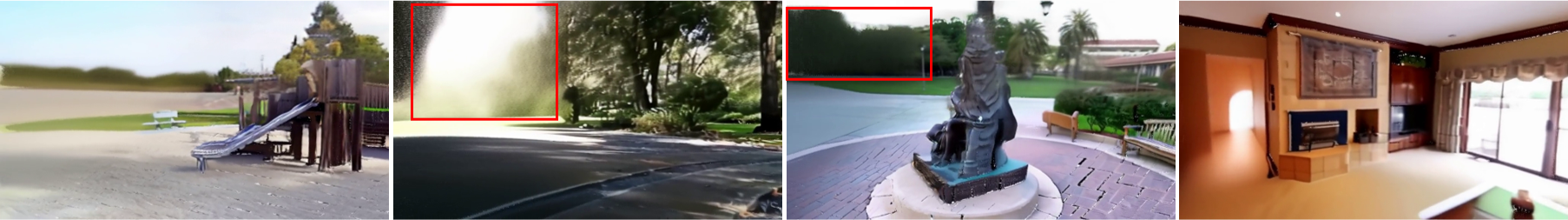}
    \end{minipage} \\
    \begin{minipage}{0.01\textwidth}
        \centering
        \small \rotatebox{90}{\textbf{Ours}}
    \end{minipage}
    &
    \begin{minipage}{1.0\textwidth}
        \centering
        \includegraphics[width=\textwidth]{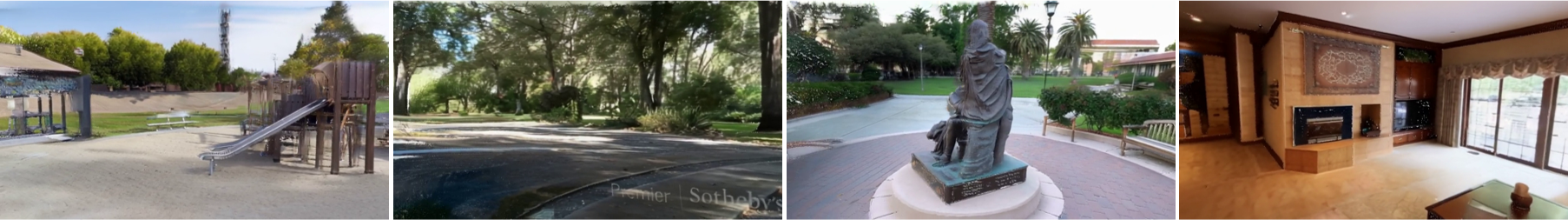}
    \end{minipage} \\
    \end{tabular}
    }
\caption{\textbf{Qualitative comparison on 3D scene generation.} We present images rendered from scenes generated by various single image-to-3D methods. The green and red bounding boxes have the same meaning as in Fig.~\ref{fig:qual-comp-novel}. Our approach achieves superior visual results.} 
\vspace{-.2cm}
\label{fig:qual-comp-scene}
\end{figure*}

\begin{table}[t]
	\centering
  \caption{\textbf{Quantitative comparison on 3D scene generation.} Scenes generated from single images by our method achieve nearly superior metric results across various datasets.}
  \label{tab:quan-comp-scene}
  \vspace{-.15cm}
         {
	\begin{tabular}{lcccccc}
		\toprule
        Dataset   & \multicolumn{3}{c}{Results on \textit{RealEstate10K}}  & \multicolumn{3}{c}{Results on \textit{Tanks and Temples}} \\ \midrule 
        Metric &  PSNR $\uparrow$ & SSIM $\uparrow$ & LPIPS $\downarrow$ & PSNR $\uparrow$ & SSIM $\uparrow$ & LPIPS $\downarrow$ \\
        \midrule
        LucidDreamer & 13.03 & 0.498 &0.590 & 11.67 &0.342 & 0.661 \\
        DimensionX &11.55 & 0.438 & 0.718 &11.02 & 0.308&0.700\\
        See3D & 14.60 & 0.544 & 0.483 &\underline{12.82} &\textbf{0.396} &0.584\\
        ViewCrafter & \underline{15.06} & \underline{0.562}& \underline{0.446} & 12.35 &0.356 & \underline{0.581} \\
        \cellcolor{gray!10}\textbf{\method} & \cellcolor{gray!10}\textbf{16.18} & \cellcolor{gray!10}\textbf{0.604}& \cellcolor{gray!10}\textbf{0.369} & \cellcolor{gray!10}\textbf{12.99} & \cellcolor{gray!10}\underline{0.389}& \cellcolor{gray!10}\textbf{0.544} \\
        \bottomrule
	\end{tabular}
    }
    \vspace{-.1in}
\end{table} 

We mainly evaluate our method for 3D scene generation by comparing the visual quality of the rendering results with 4 open-source baseline methods: LucidDreamer~\cite{lucid}, DimensionX~\cite{sun2024dimensionx}, See3D~\cite{ma2024you}, and ViewCrafter~\cite{viewcrafter}. Using the same sampling strategy as in Sec.~\ref{sec:comp-video}, we randomly selected 100 and 50 images from the RE10K~\cite{realestate10k} and Tanks~\cite{tanktemple} datasets for evaluation. Except for LucidDreamer, which generates scenes using its original implementation, scenes for other methods are reconstructed from the videos generated to 3DGS, with reconstruction hyperparameters set in~\cite{sun2024dimensionx}. We choose PSNR, SSIM, and LPIPS for the evaluation metrics to compare the renderings from the generated 3D scenes by each baseline against the ground truth frames. 

As illustrated in the qualitative comparison in Fig.~\ref{fig:qual-comp-scene}, the scenes generated by \method exhibit higher consistency with the content of the input images compared to other baselines. Furthermore, \method generates content with higher visual quality in new regions beyond the input. We also conducted a quantitative comparison, as presented in Tab.~\ref{tab:quan-comp-scene}, which shows that \method outperforms nearly all baselines in terms of metrics, with only a slight decrease compared to See3D on the SSIM in the Tanks~\cite{tanktemple} dataset. All results indicate that \method generates scenes with higher 3D consistency and visual quality.

\subsection{Ablation study}
\label{sec:abla}

\renewcommand\cellset{\renewcommand\arraystretch{1}}
\begin{figure}[th]
    \centering
    \resizebox{1.0\linewidth}{!}{
    \begin{tabular}{cc}
    &
    \begin{minipage}{1.0\linewidth}
        \centering
        \includegraphics[width=\linewidth]{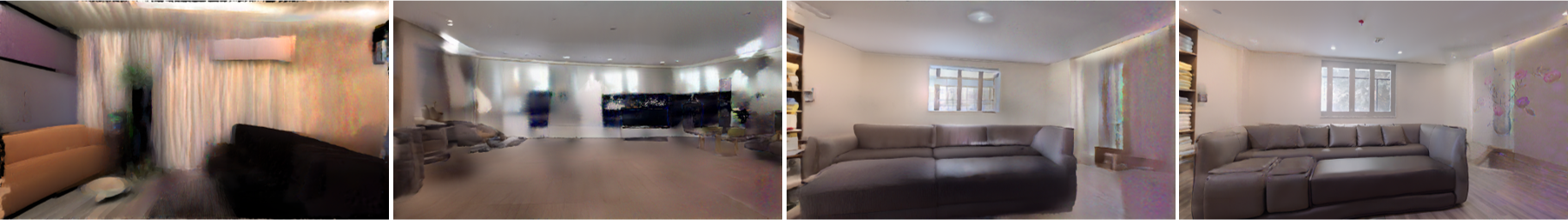}
    \end{minipage} \\
    &
    \begin{minipage}{1.0\linewidth}
        \centering
        \includegraphics[width=\linewidth]{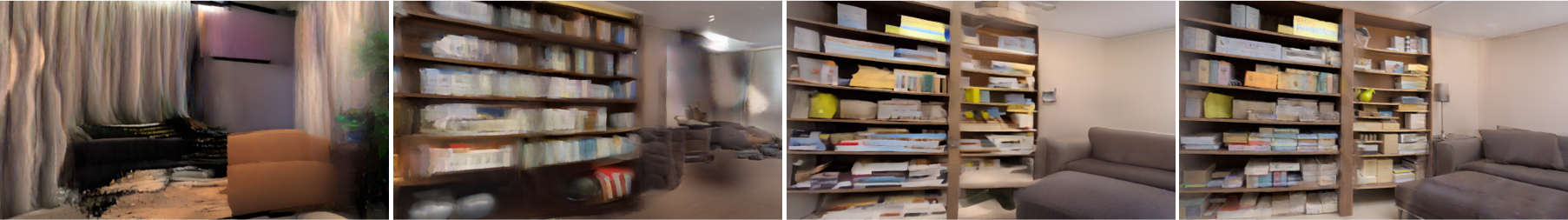}
    \end{minipage}\vspace{0.8ex} \\
    &  \begin{minipage}{1.0\linewidth}
    \makecell[cc]{
    \makebox[0.2475\linewidth]{(a) w/o V2V} \hfill
    \makebox[0.2475\linewidth]{(b) w/o zoom-out} \hfill
    \makebox[0.2475\linewidth]{(c) w/o refine} \hfill
    \makebox[0.2475\linewidth]{(d) Full}
  }\end{minipage}\\
    \end{tabular}
    }
\vspace{-.2cm}
\caption{\textbf{Ablation study.} To generate a 360° view 3D scene,  \method necessitates our video-to-video model and an initial zoom-out trajectory. Additionally, a refinement process can further enhance the visual quality of the generated 3D scene. } 
\vspace{-.2cm}
\label{fig:abla}
\end{figure}


We conduct an ablation study to demonstrate the necessity of each component in \method, as illustrated in Fig.~\ref{fig:abla}.

\textbf{Ablation on video diffusion.}
As shown in Fig.~\ref{fig:abla}a, replacing our V2V model in \method with ViewCrafter resulted in blurred scene content. This is due to inconsistencies in ViewCrafter's output under large camera variations, as discussed in Sec.~\ref{sec:3d-scene-gen}.

\textbf{Ablation on camera trajectory.} 
A zoom-out movement is crucial for enlarging the scene to enhance camera control. Without it, the generated video will mismatch with the input trajectory, leading to inconsistencies and blurriness in the generated scene, as shown in Fig.~\ref{fig:abla}b.

\textbf{Ablation on refinement process.}
The video model's generation quality restricts the detail in the generated scene. A refinement process, as discussed in Sec.~\ref{sec:3d-scene-gen}, further enhances the generated details while preserving the existing geometric structure of the scene, as shown in Fig.~\ref{fig:abla}c.


\section{Conclusion}
We present \method, a framework for generating flexible-view 3D scenes from a single image. It combines a fine-tuned video-to-video diffusion model for high-quality novel view synthesis and a progressive flexible-view 3D scene generation process. Leveraging an advanced pre-trained video foundation model and accurate training data, FlexWorld handles large camera pose variations, enabling consistent and detailed 3D scene generation that supports 360° rotations and zooming. Extensive experiments show \method's superior viewpoint flexibility and visual quality performance compared to existing methods. We believe that \method is promising and holds significant potential for virtual reality content creation and 3D tourism.


\bibliographystyle{plainnat}
\bibliography{main}

\clearpage

\beginappendix

\input{seed/appendix}

\end{document}

%% file: seed/appendix.tex

\appendix
\section{Impact of training data in video diffusion}
\label{apd:viewcrafterbad}
\renewcommand\cellset{\renewcommand\arraystretch{1}}
\begin{figure*}[ht]
    \centering
    \resizebox{1.0\textwidth}{!}{
    \begin{tabular}{cc}
    \begin{minipage}{0.01\textwidth}
        \centering
        \small \rotatebox{90}{ViewCrafter~\cite{viewcrafter}}
    \end{minipage}
    &
    \begin{minipage}{1.0\textwidth}
        \centering
        \includegraphics[width=\textwidth]{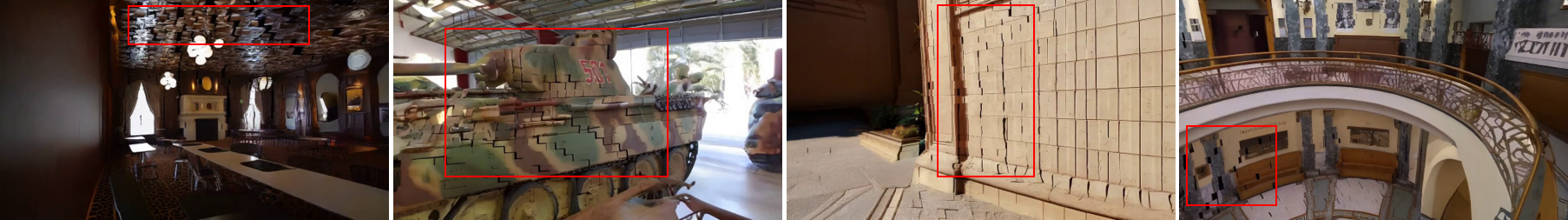}
    \end{minipage} \\
    \begin{minipage}{0.01\textwidth}
        \centering
        \small \rotatebox{90}{\textbf{Ours}}
    \end{minipage}
    &
    \begin{minipage}{1.0\textwidth}
        \centering
        \includegraphics[width=\textwidth]{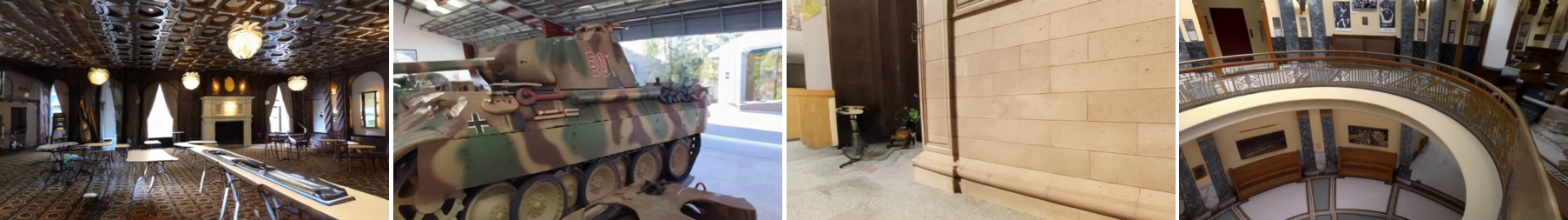}
    \end{minipage} \\
    \end{tabular}
    }
\caption{\textbf{Artifacts generated by ViewCrafter~\cite{viewcrafter}.} Compared to \method, ViewCrafter produces more artifacts that resemble those found in the incomplete videos within the training dataset constructed by its method.} 
\vspace{-.2cm}
\label{fig:viewcrafterbad}
\end{figure*}

We argue that employing a dense stereo model for data construction can introduce peculiar artifacts in the generated videos. As demonstrated in Fig.~\ref{fig:viewcrafterbad}, videos produced by ViewCrafter may exhibit artifacts closely resembling those in the incomplete data from the dataset, as shown in Fig.~\ref{fig:datacreate}. In contrast, our method generates videos free from such artifacts. This demonstrates that our dataset construction approach is more conducive to model training and better aligned with \method framework, ultimately enhancing the quality of the generated outputs.

\section{More implementation details}
\label{apd:more-detail}

\subsection{Training of V2V model.} 
\method's V2V model is based on CogVideoX-5B-I2V~\cite{yang2024cogvideox} and fine-tuned using the SAT framework. The original I2V model accepts a single image input encoded by a 3D-VAE into latents with a temporal dimension of 1, followed by zero tensor concatenation to match the dimension compressed frame. In contrast, our V2V model directly accepts video input, encoding it into latents with the compressed temporal dimension using 3D-VAE, without the need for zero tensor concatenation.

\subsection{Flexible-view 3D scene generation} 
\method begins with constructing an initial 3D point cloud from a single input image using DUSt3R~\cite{wang2024dust3r}. Since dense stereo requires paired images, we duplicate the single input image to serve as both the source and reference views. 

As for scene representation, we employ 3DGS~\cite{kerbl20233d} as the core representation, utilizing gsplat~\cite{ye2024gsplatopensourcelibrarygaussian} for implementation. When facing point cloud data (e.g., the initial point cloud), it will be immediately converted into 3DGS, serving as the initial scene representation. Unlike the original 3DGS, which uses spherical harmonics, our implementation directly represents color using RGB values. We avoid downsampling during the initialization of 3DGS from the point cloud, so the number of Gaussian counts equals the number of point clouds. Gaussian properties are initialized directly from the point cloud's position and color, with scale and opacity set to isotropic values of 3e-4 and 0.8, respectively, and rotation initialized using the identity matrix.

As for camera trajectory planning, we interpolate camera motion between the first and last frames to generate 49 camera poses, aligning with the input requirements of our V2V model. The generated video primarily utilizes linear interpolation and cubic spline interpolation in spatial coordinates, while spherical interpolation is applied to rotation matrices to ensure smooth camera transitions. To prevent collisions during camera movement, we calculate the minimum depth from the input image's estimated depth to plan the movement range.

When integrating sampled 3DGS into an existing scene, we utilize DUSt3R~\cite{wang2024dust3r} to extract consistent depth from keyframes. we select $m=6$ keyframes and employ the fully connected pairing strategy to achieve more accurate depth estimation. Keyframes are selected deterministically using a uniform sampling strategy, with the reference view typically chosen as the input image's corresponding view due to its superior visual quality and role as the starting point for scene expansion. After depth alignment, we utilize alpha maps as masks rendered from the scene to avoid the inclusion of redundant content. We apply 25 iterations of dilation to the alpha map to mitigate fragmentation in the added points.

During 3DGS training, we enhance visual quality by upscaling input video frames using the image super-resolution model Real-ESRGAN~\cite{wang2021realesrgan}. We use the original Gaussian paper's strategies for splitting, duplicating, and pruning Gaussians, but we disable the reset opacity strategy. Compared to the original Gaussian, we use higher learning rates: 1e-5 for position, 5e-3 for color, 5e-2 for opacity, 5e-4 for scale, and 1e-4 for rotation.

During the refinement process, the timestamp for the forward diffusion process is set at $0.6T$, where $T$ represents the total duration of the diffusion process. We focus on refining images when rotating cameras rather than translating ones. Specifically, we refine 5 frames from a panoramic scene using image-to-image refinement, followed by 1000 iterations of Gaussian optimization across all images to refine the overall Gaussian representation.

\section{More results}
\label{apd:more-results}
\begin{figure*}[bht]
    \centering
\resizebox{1.0\textwidth}{!}{
    \begin{tabular}{cc}
    \multicolumn{1}{c|}{\multirow{2}{*}{
    \begin{minipage}{0.2\textwidth}
        \centering
        \includegraphics[width=\textwidth]{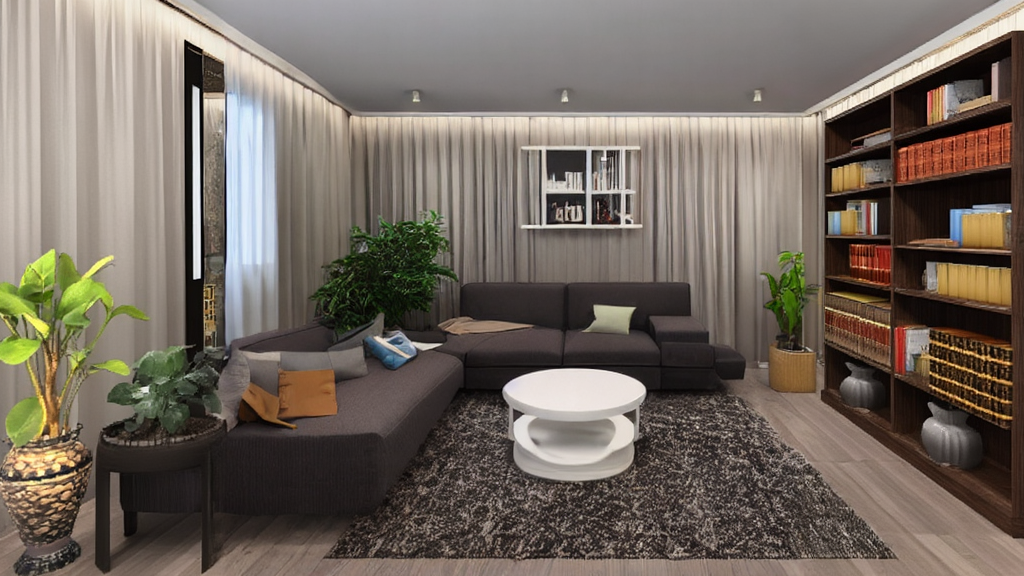}
    \end{minipage}}}
    &
    \begin{minipage}{0.8\textwidth}
        \centering
        \includegraphics[width=1.001\textwidth]{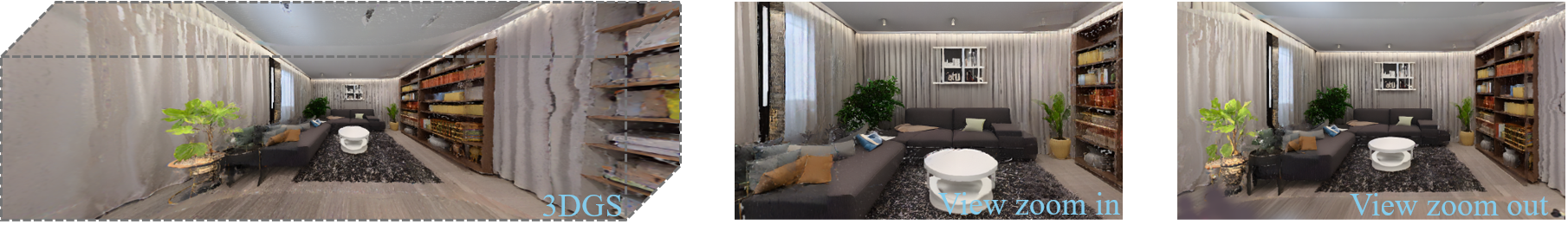}
    \end{minipage}\\
    \multicolumn{1}{c|}{}
    &
    \begin{minipage}{0.8\textwidth}
        \centering
        \includegraphics[width=\textwidth]{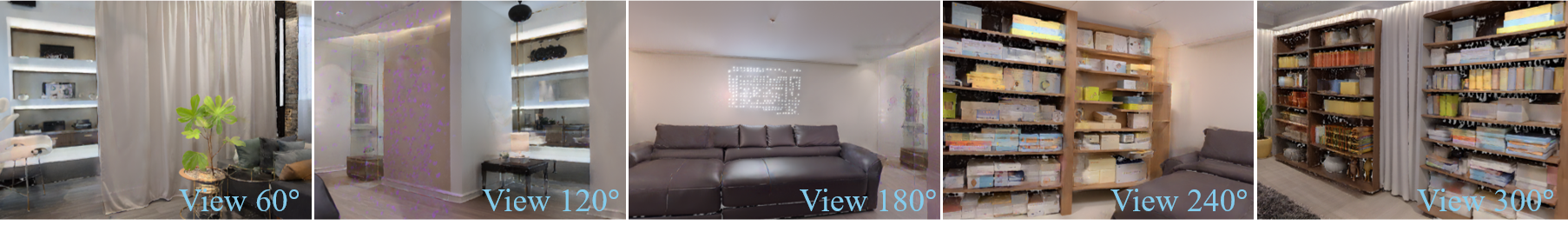}
    \end{minipage}\\
    \multicolumn{1}{c|}{\multirow{2}{*}{
    \begin{minipage}{0.2\textwidth}
        \centering
        \includegraphics[width=\textwidth]{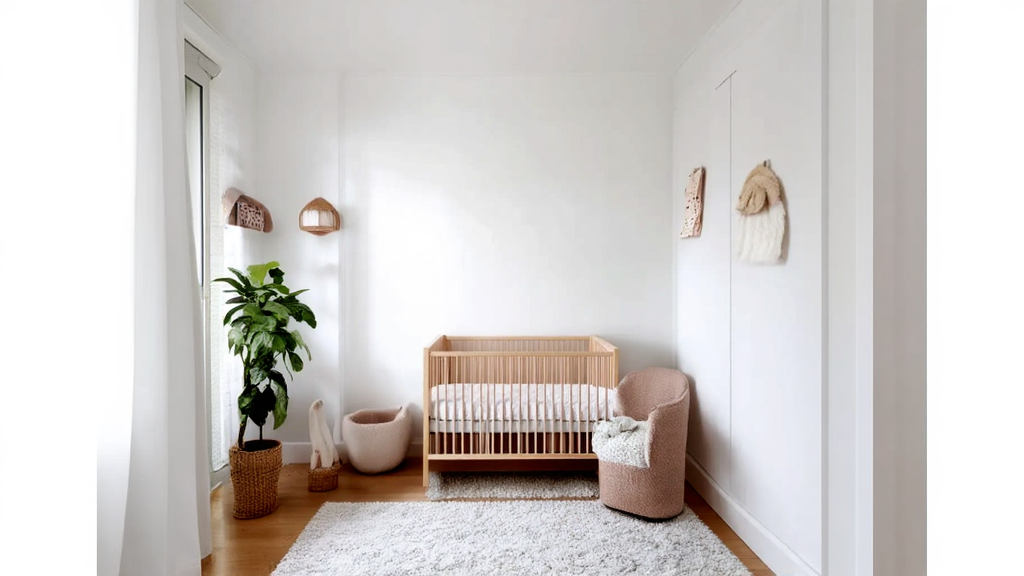}
    \end{minipage}}}
    &
    \begin{minipage}{0.8\textwidth}
        \centering
        \includegraphics[width=1.001\textwidth]{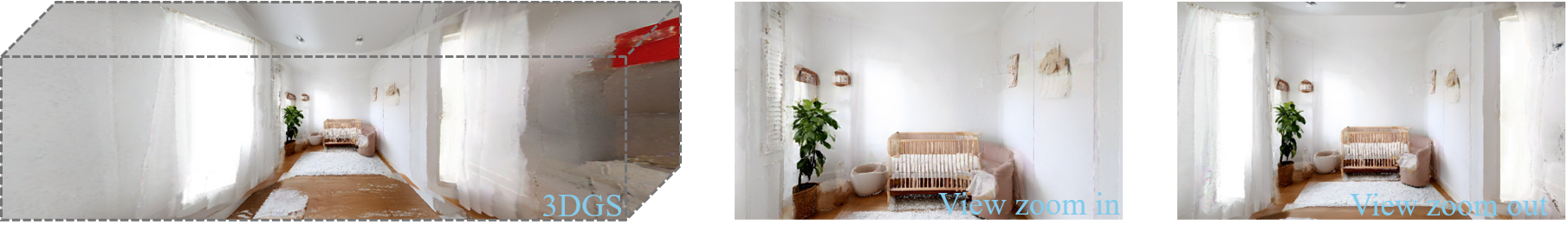}
    \end{minipage}\\
    \multicolumn{1}{c|}{}
    &
    \begin{minipage}{0.8\textwidth}
        \centering
        \includegraphics[width=\textwidth]{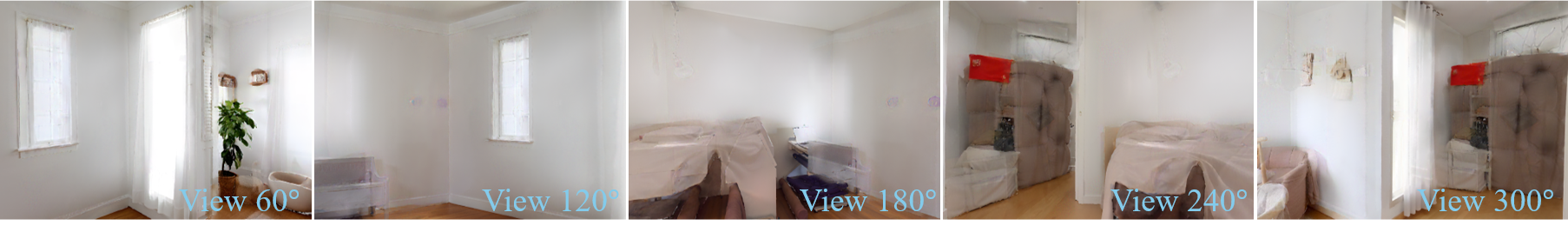}
    \end{minipage}\\

    \multicolumn{1}{c|}{\multirow{2}{*}{
    \begin{minipage}{0.2\textwidth}
        \centering
        \includegraphics[width=\textwidth]{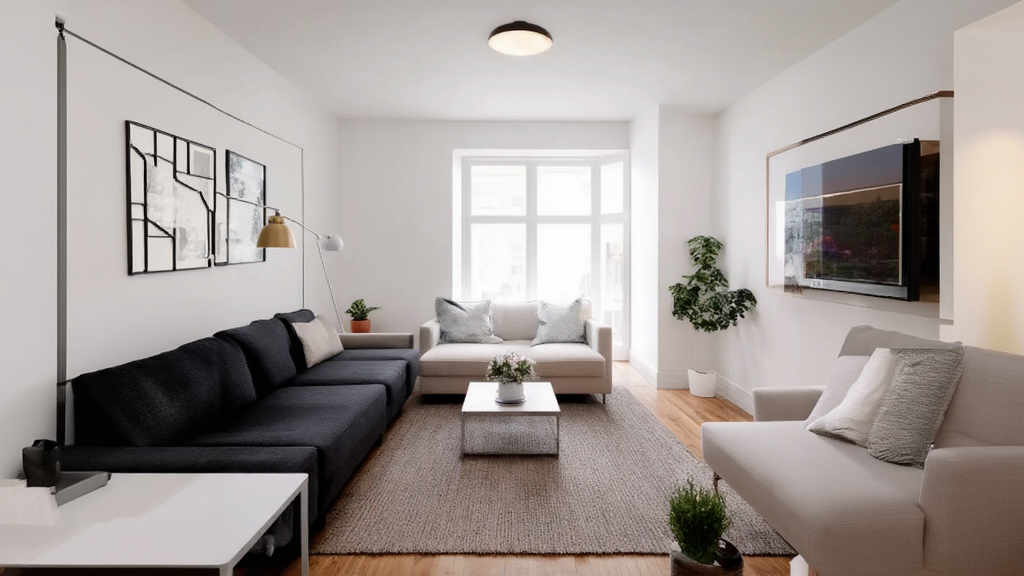}
    \end{minipage}}}
    &
    \begin{minipage}{0.8\textwidth}
        \centering
        \includegraphics[width=1.001\textwidth]{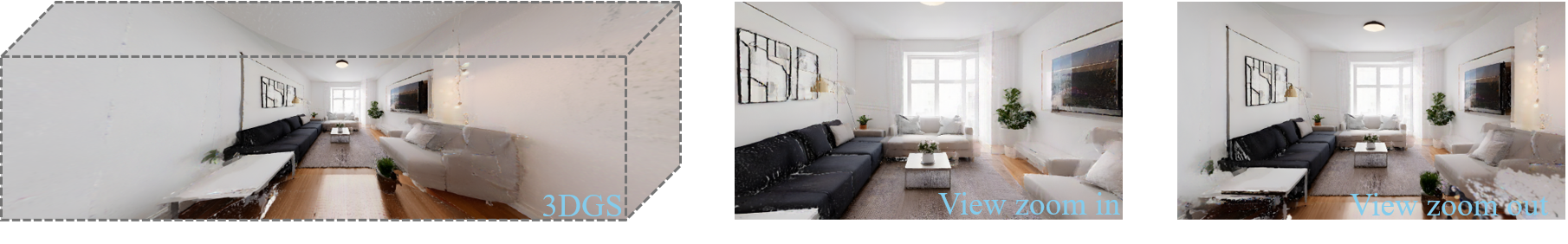}
    \end{minipage}\\
    \multicolumn{1}{c|}{}
    &
    \begin{minipage}{0.8\textwidth}
        \centering
        \includegraphics[width=\textwidth]{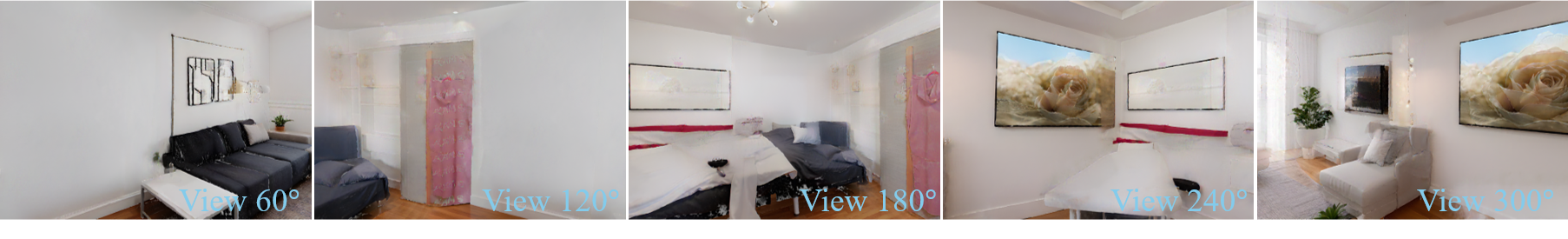}
    \end{minipage}\\
    Input &  Flexible-view 3D scene generated by \method\\
    \end{tabular}
    }
\caption{More results of generated 360° scene from \method.} 
\vspace{-.2cm}
\label{fig:more-results-scene}
\end{figure*}

\begin{figure*}[ht]
    \centering
\resizebox{1.0\textwidth}{!}{
    \begin{tabular}{cc}
    \multicolumn{1}{c|}{ 
    \begin{minipage}{0.2\textwidth}
        \centering
        \includegraphics[width=\textwidth]{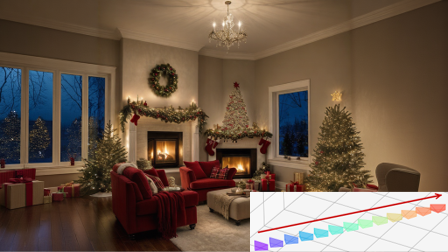}
    \end{minipage}}
    &
    \begin{minipage}{0.8\textwidth}
        \centering
        \includegraphics[width=\textwidth]{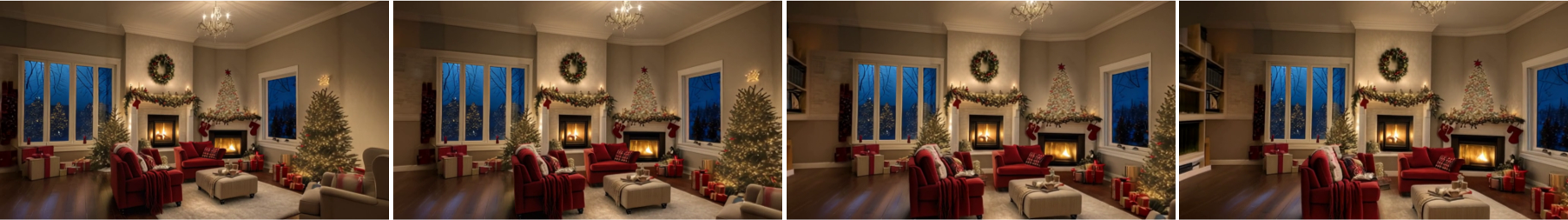}
    \end{minipage} \\
    \multicolumn{1}{c|}{
    \begin{minipage}{0.2\textwidth}
        \centering
        \includegraphics[width=\textwidth]{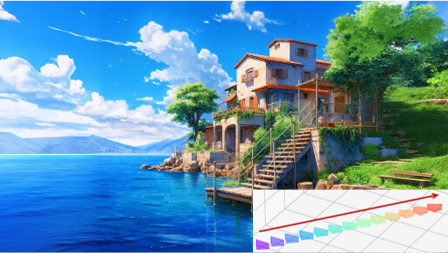}
    \end{minipage}}
    &
    \begin{minipage}{0.8\textwidth}
        \centering
        \includegraphics[width=\textwidth]{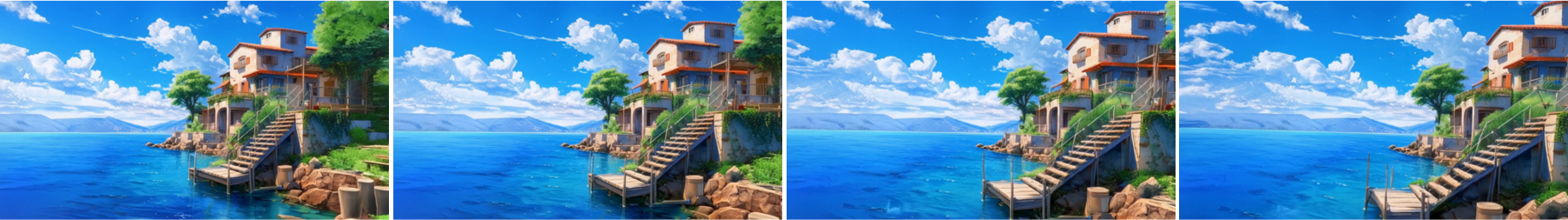}
    \end{minipage} \\
    \multicolumn{1}{c|}{
    \begin{minipage}{0.2\textwidth}
        \centering
        \includegraphics[width=\textwidth]{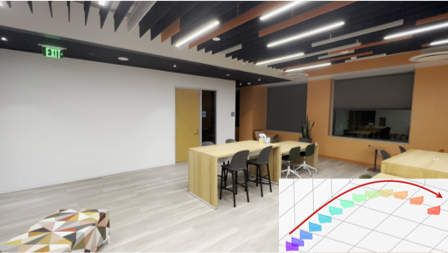}
    \end{minipage}}
    &
    \begin{minipage}{0.8\textwidth}
        \centering
        \includegraphics[width=\textwidth]{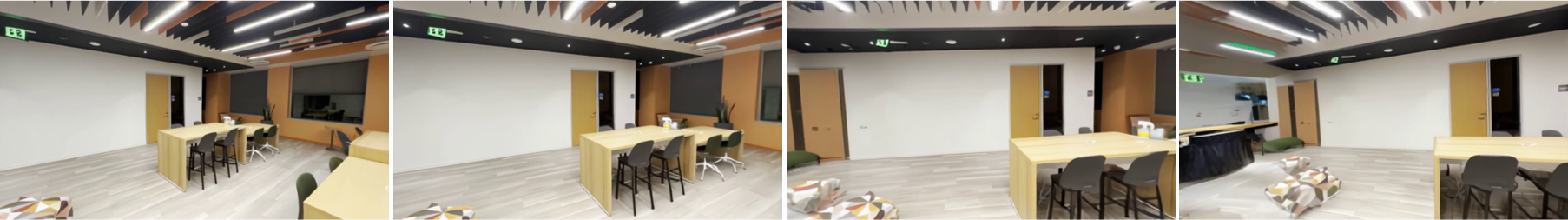}
    \end{minipage}\\
    \multicolumn{1}{c|}{
    \begin{minipage}{0.2\textwidth}
        \centering
        \includegraphics[width=\textwidth]{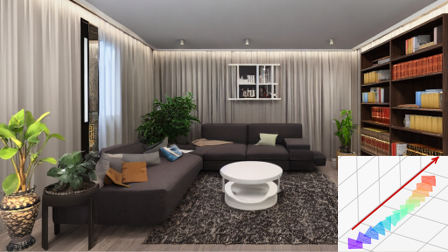}  
    \end{minipage}} 
    &
    \begin{minipage}{0.8\textwidth}
        \centering
        \includegraphics[width=\textwidth]{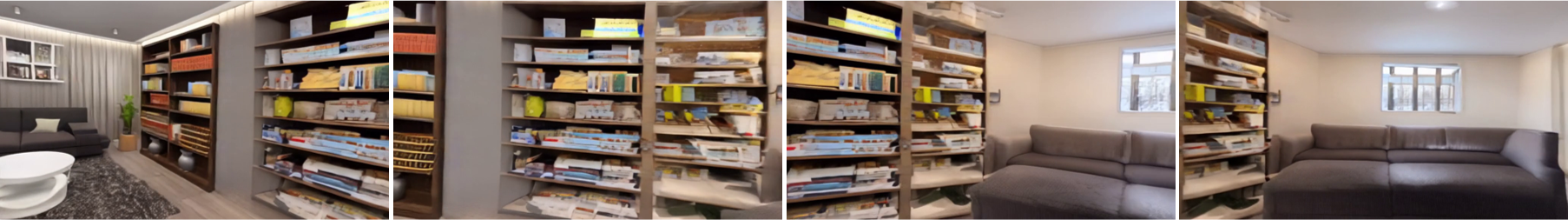} 
    \end{minipage} \\ 
    \multicolumn{1}{c|}{
    \begin{minipage}{0.2\textwidth}
        \centering
        \includegraphics[width=\textwidth]{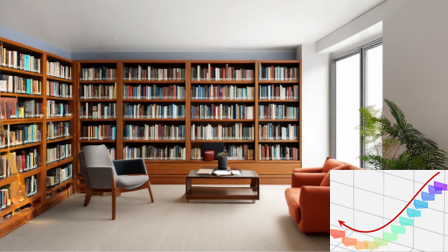}  
    \end{minipage}} 
    &
    \begin{minipage}{0.8\textwidth}
        \centering
        \includegraphics[width=\textwidth]{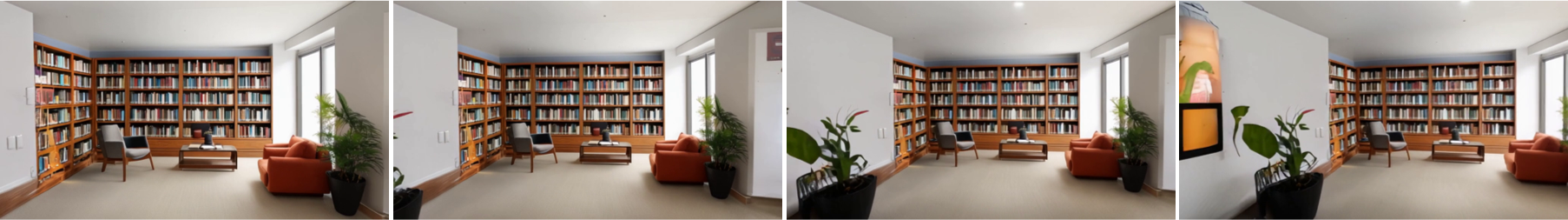} 
    \end{minipage} \\ 
    \multicolumn{1}{c|}{
    \begin{minipage}{0.2\textwidth}
        \centering
        \includegraphics[width=\textwidth]{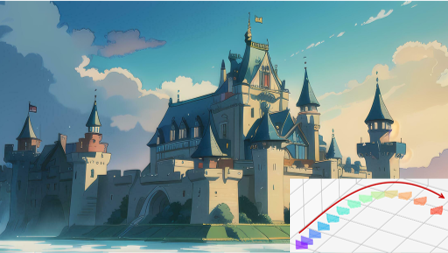}  
    \end{minipage}} 
    &
    \begin{minipage}{0.8\textwidth}
        \centering
        \includegraphics[width=\textwidth]{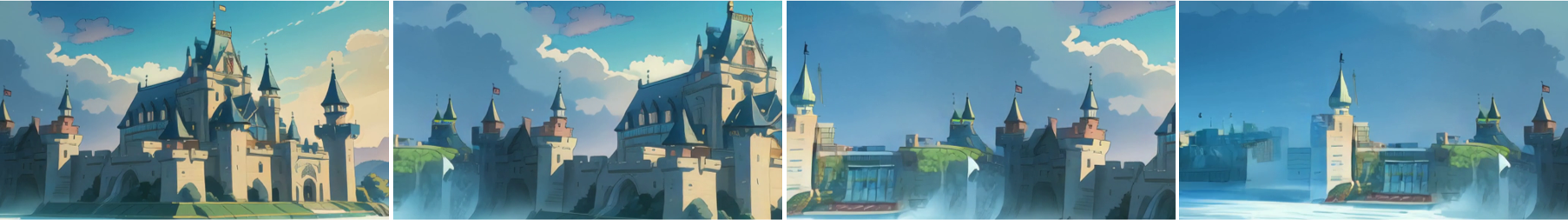} 
    \end{minipage} \\ 
    \multicolumn{1}{c|}{
    \begin{minipage}{0.2\textwidth}
        \centering
        \includegraphics[width=\textwidth]{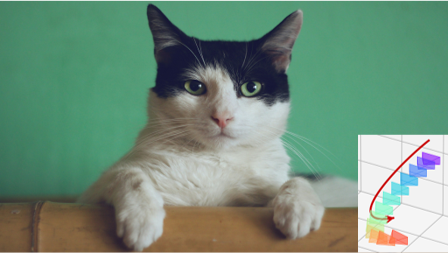}  
    \end{minipage}} 
    &
    \begin{minipage}{0.8\textwidth}
        \centering
        \includegraphics[width=\textwidth]{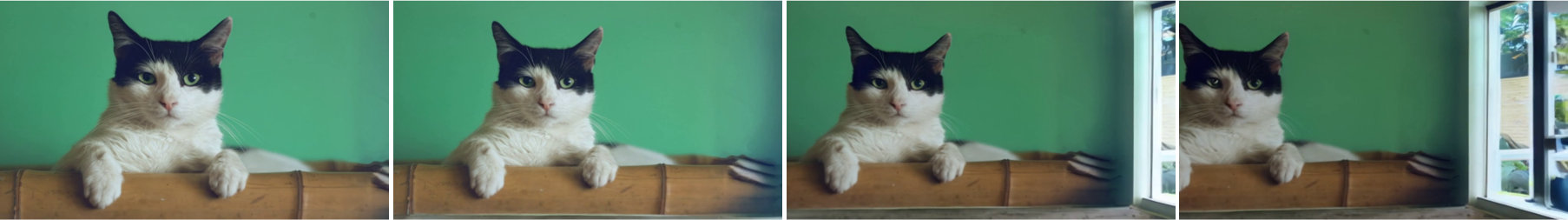} 
    \end{minipage} \\ 
    \multicolumn{1}{c|}{
    \begin{minipage}{0.2\textwidth}
        \centering
        \includegraphics[width=\textwidth]{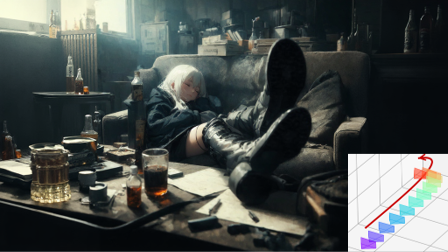}  
    \end{minipage}} 
    &
    \begin{minipage}{0.8\textwidth}
        \centering
        \includegraphics[width=\textwidth]{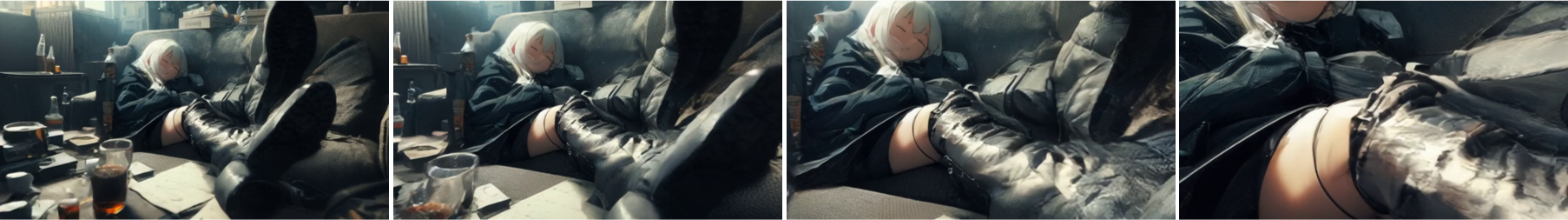} 
    \end{minipage} \\ 
    \multicolumn{1}{c|}{
    \begin{minipage}{0.2\textwidth}
        \centering
        \includegraphics[width=\textwidth]{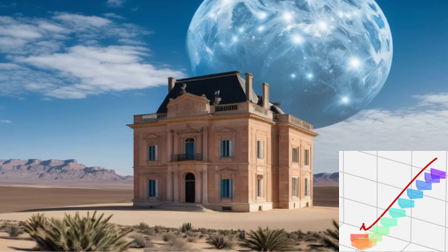}  
    \end{minipage}} 
    &
    \begin{minipage}{0.8\textwidth}
        \centering
        \includegraphics[width=\textwidth]{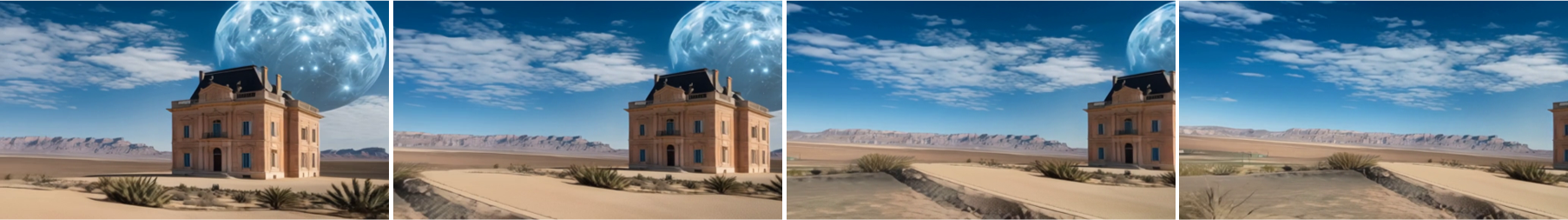} 
    \end{minipage} \\ 
    \multicolumn{1}{c|}{
    \begin{minipage}{0.2\textwidth}
        \centering
        \includegraphics[width=\textwidth]{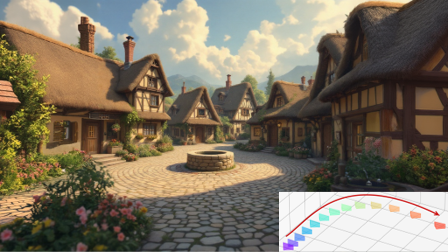}  
    \end{minipage}} 
    &
    \begin{minipage}{0.8\textwidth}
        \centering
        \includegraphics[width=\textwidth]{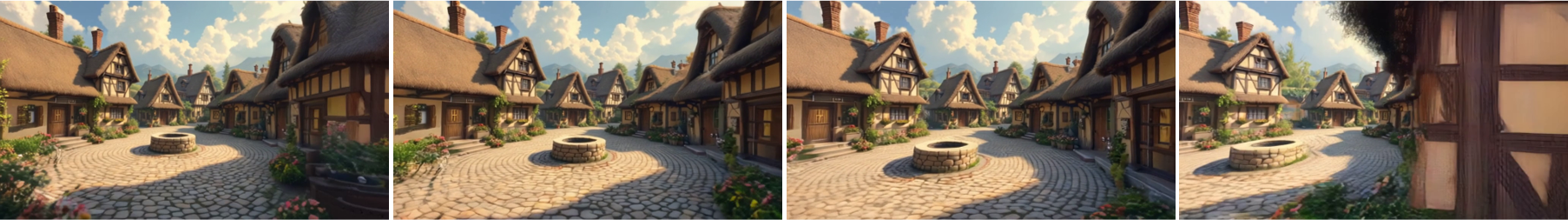} 
    \end{minipage} \\ 
    \multicolumn{1}{c|}{
    \begin{minipage}{0.2\textwidth}
        \centering
        \includegraphics[width=\textwidth]{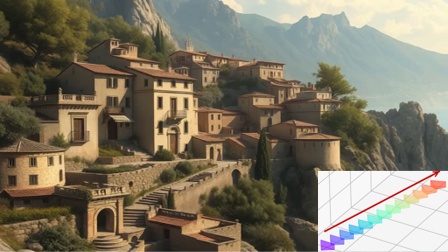}  
    \end{minipage}} 
    &
    \begin{minipage}{0.8\textwidth}
        \centering
        \includegraphics[width=\textwidth]{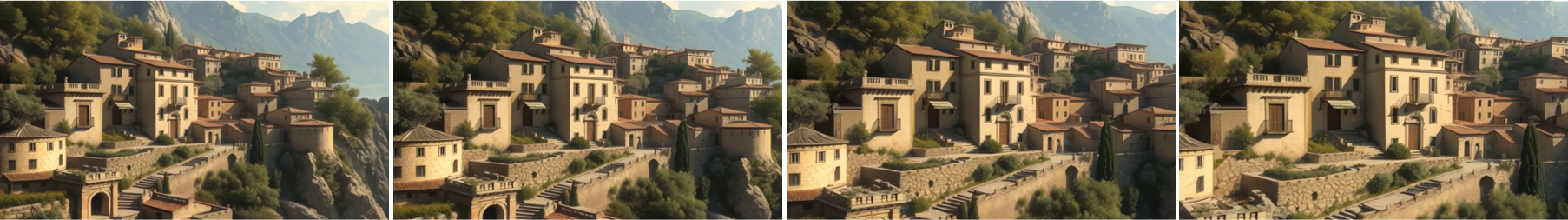} 
    \end{minipage} \\ 
    Input & Videos generated by our V2V model given camera trajectories \vspace{0.8ex}\\
    \end{tabular}
    }
\caption{More results of generated videos from \method.} 
\vspace{-.2cm}
\label{fig:more-results-video}
\end{figure*}

We present more results for our video generation in Fig.~\ref{fig:more-results-video} and the 360° 3D scene generation in Fig.~\ref{fig:more-results-scene}.

\renewcommand\cellset{\renewcommand\arraystretch{1}}

\renewcommand\cellset{\renewcommand\arraystretch{1}}
